\documentclass[letterpaper, 10 pt, conference]{ieeeconf}  % Comment this line out if you need a4paper

\IEEEoverridecommandlockouts                              % This command is only needed if 
\usepackage{graphicx} % for pdf, bitmapped graphics files

\usepackage{comment}
\usepackage{lipsum}
\usepackage{booktabs}
\usepackage[table]{xcolor}
\usepackage{pifont}
\usepackage{bbm}
\usepackage{multirow}

\usepackage{dsfont}
\usepackage[absolute,overlay]{textpos}
\usepackage{amsmath} % assumes amsmath package installed
\usepackage{amssymb}  % assumes amsmath package installed
\usepackage[hidelinks]{hyperref}

\title{\LARGE \bf
Instance--Guided Unsupervised Domain Adaptation for Robotic Semantic Segmentation 
}
\author{Michele Antonazzi, Lorenzo Signorelli, Matteo Luperto, Nicola Basilico
\thanks{\textcolor{black}{All authors are with the Department of Computer Science, University of
Milan, Milano, Italy. Email \texttt{name.surname@unimi.it}}
}}

\begin{document}

\begin{textblock*}{10cm}(1.5cm,1cm) % {block width} (coords) 
   Accepted for publication at ICRA 2026
\end{textblock*}

\maketitle
\thispagestyle{empty}
\pagestyle{empty}

\newcommand{\imageone}[0]{0}
\newcommand{\imagetwo}[0]{1}
\newcommand{\imagethree}[0]{1}
\newcommand{\imagefour}[0]{1}
\newcommand{\imagefive}[0]{1}
\newcommand{\bestvalue}[1]{\textbf{#1}}
\newcommand{\secondvalue}[1]{\underline{#1}}

\newcommand{\ww}[0]{0.175\linewidth}
\newcommand{\IRI}[0]{IR$_\texttt{I}$}
\newcommand{\IRG}[0]{IR$_\texttt{G}$}
\newcommand{\IRGI}[0]{IR$_\texttt{GI}$}
\newcommand{\MC}[0]{MC}
\newcommand{\PT}[0]{PT}
\definecolor{color_wall}{RGB}{174, 199, 232}
\definecolor{color_floor}{RGB}{152, 223, 138}
\definecolor{color_cabinet}{RGB}{31, 119, 180}
\definecolor{color_bed}{RGB}{255, 187, 120}
\definecolor{color_chair}{RGB}{188, 189, 34}
\definecolor{color_sofa}{RGB}{140, 86, 75}
\definecolor{color_table}{RGB}{255, 152, 150}
\definecolor{color_door}{RGB}{214, 39, 40}
\definecolor{color_window}{RGB}{197, 176, 213}
\definecolor{color_bookshelf}{RGB}{148, 103, 189}
\definecolor{color_picture}{RGB}{196, 156, 148}
\definecolor{color_counter}{RGB}{23, 190, 207}
\definecolor{color_blinds}{RGB}{178, 76, 76}
\definecolor{color_desk}{RGB}{247, 182, 210}
\definecolor{color_shelves}{RGB}{66, 188, 102}
\definecolor{color_curtain}{RGB}{219, 219, 141}
\definecolor{color_dresser}{RGB}{140, 57, 197}
\definecolor{color_pillow}{RGB}{202, 185, 52}
\definecolor{color_mirror}{RGB}{51, 176, 203}
\definecolor{color_floormat}{RGB}{200, 54, 131}
\definecolor{color_clothes}{RGB}{92, 193, 61}
\definecolor{color_ceiling}{RGB}{78, 71, 183}
\definecolor{color_books}{RGB}{172, 114, 82}
\definecolor{color_refrigerator}{RGB}{255, 127, 14}
\definecolor{color_television}{RGB}{91, 163, 138}
\definecolor{color_paper}{RGB}{153, 98, 156}
\definecolor{color_towel}{RGB}{140, 153, 101}
\definecolor{color_showercurtain}{RGB}{158, 218, 229}
\definecolor{color_box}{RGB}{100, 125, 154}
\definecolor{color_whiteboard}{RGB}{178, 127, 135}
\definecolor{color_person}{RGB}{120, 185, 128}
\definecolor{color_nightstand}{RGB}{146, 111, 194}
\definecolor{color_toilet}{RGB}{44, 160, 44}
\definecolor{color_sink}{RGB}{112, 128, 144}
\definecolor{color_lamp}{RGB}{96, 207, 209}
\definecolor{color_bathtub}{RGB}{227, 119, 194}
\definecolor{color_bag}{RGB}{213, 92, 176}
\definecolor{color_otherstructure}{RGB}{94, 106, 211}
\definecolor{color_otherfurniture}{RGB}{82, 84, 163}
\definecolor{color_otherprop}{RGB}{100, 85, 144}

\newcommand{\classwall}{\textcolor{color_wall}{\ding{108}}~\texttt{wall}}
\newcommand{\classfloor}{\textcolor{color_floor}{\ding{108}}~\texttt{floor}}
\newcommand{\classcabinet}{\textcolor{color_cabinet}{\ding{108}}~\texttt{cabinet}}
\newcommand{\classbed}{\textcolor{color_bed}{\ding{108}}~\texttt{bed}}
\newcommand{\classchair}{\textcolor{color_chair}{\ding{108}}~\texttt{chair}}
\newcommand{\classsofa}{\textcolor{color_sofa}{\ding{108}}~\texttt{sofa}}
\newcommand{\classtable}{\textcolor{color_table}{\ding{108}}~\texttt{table}}
\newcommand{\classdoor}{\textcolor{color_door}{\ding{108}}~\texttt{door}}
\newcommand{\classwindow}{\textcolor{color_window}{\ding{108}}~\texttt{window}}
\newcommand{\classbookshelf}{\textcolor{color_bookshelf}{\ding{108}}~\texttt{bookshelf}}
\newcommand{\classpicture}{\textcolor{color_picture}{\ding{108}}~\texttt{picture}}
\newcommand{\classcounter}{\textcolor{color_counter}{\ding{108}}~\texttt{counter}}
\newcommand{\classblinds}{\textcolor{color_blinds}{\ding{108}}~\texttt{blinds}}
\newcommand{\classdesk}{\textcolor{color_desk}{\ding{108}}~\texttt{desk}}
\newcommand{\classshelves}{\textcolor{color_shelves}{\ding{108}}~\texttt{shelves}}
\newcommand{\classcurtain}{\textcolor{color_curtain}{\ding{108}}~\texttt{curtain}}
\newcommand{\classdresser}{\textcolor{color_dresser}{\ding{108}}~\texttt{dresser}}
\newcommand{\classpillow}{\textcolor{color_pillow}{\ding{108}}~\texttt{pillow}}
\newcommand{\classmirror}{\textcolor{color_mirror}{\ding{108}}~\texttt{mirror}}
\newcommand{\classfloormat}{\textcolor{color_floormat}{\ding{108}}~\texttt{floormat}}
\newcommand{\classclothes}{\textcolor{color_clothes}{\ding{108}}~\texttt{clothes}}
\newcommand{\classceiling}{\textcolor{color_ceiling}{\ding{108}}~\texttt{ceiling}}
\newcommand{\classbooks}{\textcolor{color_books}{\ding{108}}~\texttt{books}}
\newcommand{\classrefrigerator}{\textcolor{color_refrigerator}{\ding{108}}~\texttt{refrigerator}}
\newcommand{\classtelevision}{\textcolor{color_television}{\ding{108}}~\texttt{television}}
\newcommand{\classpaper}{\textcolor{color_paper}{\ding{108}}~\texttt{paper}}
\newcommand{\classtowel}{\textcolor{color_towel}{\ding{108}}~\texttt{towel}}
\newcommand{\classshowercurtain}{\textcolor{color_showercurtain}{\ding{108}}~\texttt{showercurtain}}
\newcommand{\classbox}{\textcolor{color_box}{\ding{108}}~\texttt{box}}
\newcommand{\classwhiteboard}{\textcolor{color_whiteboard}{\ding{108}}~\texttt{whiteboard}}
\newcommand{\classperson}{\textcolor{color_person}{\ding{108}}~\texttt{person}}
\newcommand{\classnightstand}{\textcolor{color_nightstand}{\ding{108}}~\texttt{nightstand}}
\newcommand{\classtoilet}{\textcolor{color_toilet}{\ding{108}}~\texttt{toilet}}
\newcommand{\classsink}{\textcolor{color_sink}{\ding{108}}~\texttt{sink}}
\newcommand{\classlamp}{\textcolor{color_lamp}{\ding{108}}~\texttt{lamp}}
\newcommand{\classbathtub}{\textcolor{color_bathtub}{\ding{108}}~\texttt{bathtub}}
\newcommand{\classbag}{\textcolor{color_bag}{\ding{108}}~\texttt{bag}}
\newcommand{\classotherstructure}{\textcolor{color_otherstructure}{\ding{108}}~\texttt{otherstructure}}
\newcommand{\classotherfurniture}{\textcolor{color_otherfurniture}{\ding{108}}~\texttt{furniture}}
\newcommand{\classotherprop}{\textcolor{color_otherprop}{\ding{108}}~\texttt{object}}

\begin{abstract}
Semantic segmentation networks, which are essential for robotic perception, often suffer from performance degradation when the visual distribution of the deployment environment differs from that of the source dataset on which they were trained. Unsupervised Domain Adaptation (UDA) addresses this challenge by adapting the network to the robot’s target environment without external supervision, leveraging the large amounts of data a robot might naturally collect during long--term operation. In such settings, UDA methods can exploit multi--view consistency across the environment’s map to fine--tune the model in an unsupervised fashion and mitigate domain shift. However, these approaches remain sensitive to cross--view instance--level inconsistencies. In this work, we propose a method\footnote{ \href{https://aislab.di.unimi.it/research/instanceuda}{https://aislab.di.unimi.it/research/instanceuda}} that starts from a volumetric 3D map to generate multi--view consistent pseudo--labels. We then refine these labels using the zero--shot instance segmentation capabilities of a foundation model, enforcing instance--level coherence. The refined annotations serve as supervision for self--supervised fine--tuning, enabling the robot to adapt its perception system at deployment time. Experiments on real--world data demonstrate that our approach consistently improves performance over state--of--the--art UDA baselines based on multi--view consistency, without requiring any ground--truth labels in the target domain.
\end{abstract}
\section{Introduction}\label{sec:introduction}
Semantic segmentation is a key component of robotic vision, as it enables autonomous systems to interpret complex environments and reason about object categories at the pixel level. This capability can boost core robotic tasks such as navigation, planning, and interaction~\cite{semantic_segmentation_survey}. Typically, segmentation models are trained on source datasets collected under controlled conditions, but their performance degrades when deployed in target environments characterized by different, though related, data distributions~\cite{wilson2020survey}. This phenomenon, known as \emph{domain shift}, is particularly critical for mobile robots, which must operate for long periods in previously unseen environments where annotated data are unavailable at training time and prohibitively expensive to acquire during deployment. A common strategy to mitigate domain shift is to collect and manually annotate target data for fine--tuning~\cite{antonazzi2025jfr}, but this process is costly, time--consuming, and often impractical in real robotic deployments~\cite{scannet}.

\begin{figure}[!t]
\centering
\begin{tabular}{@{}c@{ }c@{ }c@{ }}
{\small Robot perception} & {\small View consistency~\cite{continual_adaptation_2d3d}} & {\small Our method} \\
\includegraphics[width=0.32\linewidth]{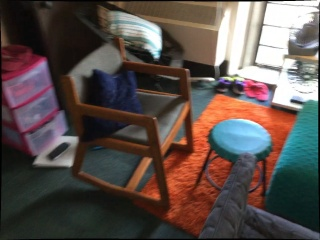}
    &
    \includegraphics[width=0.32\linewidth]{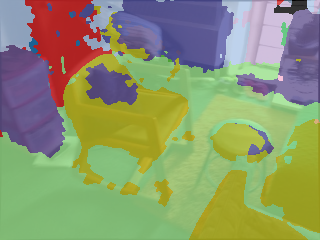}
    &
    \includegraphics[width=0.32\linewidth]{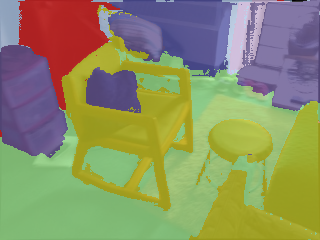}
 \\
 \includegraphics[width=0.32\linewidth]{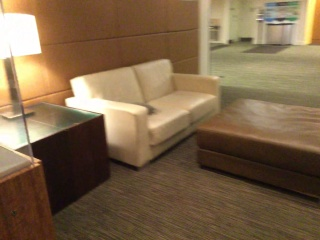}
    &
    \includegraphics[width=0.32\linewidth]{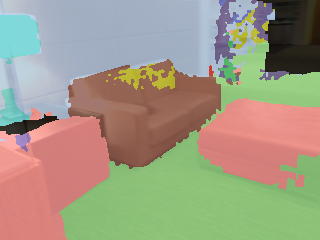}
    &
    \includegraphics[width=0.32\linewidth]{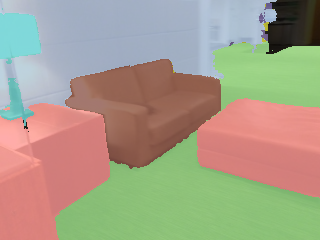}
 \\
% \includegraphics[width=0.32\linewidth]{contents/images/pseudo_labels/3cm/informed/rgb_\imagethree}
    %&
    %\includegraphics[width=0.32\linewidth]{contents/images/pseudo_labels/3cm/informed/kimera_\imagethree}
    %&
    %\includegraphics[width=0.32\linewidth]{contents/images/pseudo_labels/3cm/informed/sam_\imagethree}
% \\
 
\end{tabular}
    \caption{Examples of the improvement provided by our method (third column) to the view--consistent pseudo--labels~\cite{continual_adaptation_2d3d} (middle column).  }
    \label{fig:proposal_selection_example}
    \vspace{-0.6cm}
\end{figure}

Unsupervised Domain Adaptation (UDA) addresses the challenge of adapting a model to a new environment without relying on ground--truth labels or humans in the loop~\cite{domain_adaptation_survey}. With mobile robots, UDA can leverage domain--specific opportunities: unlike with static computer vision benchmarks, a robot is continuously immersed in its target environment, can acquire multiple observations of the same scene, and can exploit mapping and view consistency as natural proxies for model adaptation. Previous works~\cite{continual_adaptation_2d3d, informative_planning} have shown that aggregating predictions across viewpoints in a 3D environmental representation~\cite{kimera} improves pseudo--labels by enforcing spatial consistency, yet such methods remain vulnerable to instance--level incoherence that might persist across frames. As shown in Fig.~\ref{fig:proposal_selection_example}, the view--consistent pseudo--labels (middle column) obtained from the 3D map present visual artifacts generated by the rendering process and, more importantly, fail to be instance--coherent as multiple categories are assigned to the same object.

\renewcommand{\imageone}{4.png}
\renewcommand{\imageone}{4.png}
\renewcommand{\imagetwo}{11.png}
\renewcommand{\imagethree}{14.png}

In this work, we tackle and mitigate these issues by proposing an instance--guided approach to UDA for robotic semantic segmentation. Starting from multi--view consistent pseudo--labels, we integrate a refinement stage that leverages the zero--shot instance segmentation capabilities of a foundation model, queried via two different and complementary automated prompting strategies. Our method enforces instance--level coherence and mitigates rendering artifacts, yielding more accurate and stable annotations for self--supervised fine--tuning (see Fig.~\ref{fig:proposal_selection_example}, third column). Experiments on real--world data show that our method significantly outperforms existing UDA baselines, while requiring no ground--truth labels in the target domain.

In summary, this work's contributions are: (i) we propose a novel framework for UDA in robotic semantic segmentation that combines multi--view consistency with instance--aware pseudo--label refinement; (ii) we integrate the SAM foundation model into the refinement stage, devising two automatic and complementary prompting strategies; (iii) we carry out an extensive experimental evaluation on real--world data assessing how our method improves adaptation performance over existing baselines both in terms of pseudo--label quality and segmentation performance.  

\section{Related Works}

%\subsection{UDA Approaches for Semantic Segmentation}
\noindent
\textbf{UDA approaches for semantic segmentation.} Domain shift is caused by a difference in the distribution of the dataset collected in the environments used for training and evaluation.
One of the mainstream approaches for unsupervised domain adaptation in semantic segmentation tasks leverages adversarial learning to align the distributions of train and test domains used at input, intermediate features, and/or output levels. The pioneer work of~\cite{image_to_image_translation} presents a general paradigm based on unpaired image--to--image translation to align the neural network's embeddings extracted from images from different domains. Each training image is translated from its source domain to the other, thus building a latent space in which the distribution of image features is domain invariant, and semantically similar images are projected into close vicinity. %Using them, the proposed method leverages adversarial learning and cycle consistency to build a latent space for the two domains in which feature distributions are domain invariant, and semantically similar images are projected into close vicinity. 
This paradigm is extended by CrDoCo~\cite{crdoco}, which combines massive image--to--image translation using GANs with adversarial learning and pixel--wise consistency. %The goal is to make the features extracted from an image and its restyled version indistinguishable, while ensuring that their predictions remain the same.

%Instead of concentrating on inputs and inner features of the neural network, 
Another promising solution is to perform adversarial learning on the neural network's output space. %The rationale is that the segmentation labels coming from different domains should be similar in terms of global layout and local context.  
The rationale behind this is that the spatial distribution of objects in the images from the training and evaluation datasets should be similar, even if their visual features differ; an example of this is the fact that tables and chairs are usually located close to each other. Consequently, when correctly identified, the spatial distribution of labels coming from different domains should be similar.
Following this, the authors of~\cite{adapt_stuctured_output} propose to train a CNN to fool a discriminator tasked to disambiguate the domain from which its output masks are generated.
%An extension of this approach is ADVENT~\cite{advent_entropy}, a method that performs UDA leveraging the predictions' entropy, calculated using the discrete pixel--wise distribution probability over object categories. Since the entropy is usually low on data from the source domain, and high on that from the target one, the authors propose an adversarial training scheme to make the target's entropy distribution similar to that of the source.
An extension of this approach is ADVENT~\cite{advent_entropy}, which employs an adversarial training scheme to align the entropy distribution of the target images (computed as the pixel--wise categorical probability distribution) with that of the source domain, under the assumption that the entropy is usually low on data from the source domain and high on that from the target one.

%The adversarial learning paradigm has some limitations: it is inherently unstable and, by focusing only on domain confusion, completely ignores domain--specific cues. To overcome these issues, recent works leverage self--supervision with pseudo--labels inferred by the model itself from the target data. Since inference on the target domain is inherently noisy and error--prone, the naive approach of using directly pseudo--labels to finetune a mode is impractical, due to the high risk for the model of overfitting its own misclassifications.
The adversarial learning paradigm has some limitations: it is inherently unstable, it focuses only on domain confusion, and it completely ignores domain--specific cues. To overcome these issues, recent works leverage self--supervision with pseudo--labels inferred by the model itself from the target data.  However, pseudo--labels from images in the target domain are noisy and inaccurate, and they cannot be directly used to finetune the source model.
%Since inference on the target domain is inherently noisy and error--prone, the naive approach of using directly pseudo--labels to finetune a mode is impractical, due to the high risk for the model of overfitting its own misclassifications.
A promising solution to this is to balance the influence of pseudo--labels during training, promoting high--confident predictions while penalizing noisy annotations~\cite{uncertanty_pseudo_labels}. The confidence is estimated as the divergence between the outputs from two independent prediction layers. This uncertainty is then used as a regularization term during fine--tuning. 
An alternative approach is that of~\cite{prototypical_labels_denoising}, which estimates the confidence of pseudo--labels by using the relative distance of the extracted features from their respective class centroid. %The prototype for each object category is efficiently computed online during training (using the moving average of the cluster centroids inside mini--batches) to avoid the early overfitting on noisy predictions.
%A promising solution to this is to balance the influence of pseudo--labels during training, promoting high--confident predictions while penalizing noisy annotations; the work of~\cite{uncertanty_pseudo_labels} do so by estimating the uncertainty of the predictions at a pixel level by computing the divergence between the outputs from two independent prediction layers. This uncertainty is then used as a regularization term during fine--tuning to rectify the learning from noisy labels. 
%An alternative approach is those of~\cite{prototypical_labels_denoising}, where the weighting of pseudo--labels is calculated using the relative distance of the extracted features from their respective class centroid. The prototype for each object category is efficiently computed online during training (using the moving average of the cluster centroids inside mini--batches) to avoid the early overfitting on noisy predictions. 

Another general approach is to use a student network, tasked with adapting to the target domain, supervised by a teacher network that, being gradually updated using an exponential moving average of the student, provides more stable and consistent pseudo--labels. Following this, the work in~\cite{class_balanced_self_training} rectifies the pseudo--labels using feature clustering on pixel embeddings from the target domain. %A teacher network produces the embeddings for each pixel that are clustered together with a batch of randomly sampled features. 
The label of each pixel is updated based on the assigned cluster, thus providing more coherent and stable annotations that are used to supervise the student training.
A similar approach is to use Masked Image Consistency~\cite{mic}, in which the student has access to images in which random patches are masked, and it should be consistent with the teacher network, which uses unmasked ones.
%The authors of~\cite{mic} propose Masked Image Consistency (MIC), a technique in which the student is tasked to produce pseudo--labels on images in which random patches are masked. %By doing this, the student (which is trained to be consistent with a teacher that access the unmasked data), learns to predict the class of missing portions using the global context of the image, improving generalization on the unseen domain. 
%The student, what has access to masked data, is asked to be consistent with the teacher network, who has access to unmasked data; this allows the model to learn how to predict the class of missing portions, thus improving its generalisation capabilities in unseen domains.   
An extension of the teacher--student paradigm is presented in~\cite{revisiting_uda}. In such a work, a student network is supervised by two teachers: the first one is applied to the outputs while the second, bigger teacher is fixed during training and provides transfer knowledge in the feature space.

%Originally devised for sim--to--real transfer in autonomous driving benchmarks, the aforementioned approaches operate in a single--frame manner, producing independent predictions for each image. 

\noindent
\textbf{Leveraging robot embodiment for UDA.} One major limitation of the previously presented approaches is that they operate by processing each frame independently, thus neglecting possible relations with previous and past perceptions. This ignores the fact that indoor mobile robots, while navigating,  observe several similar views from the same scene; thus, the images in the sequence are highly correlated.  This fact can be leveraged as a natural proxy for consistency, which can be exploited to improve pseudo--label quality, reduce uncertainty, and enhance the recognition of small, occluded, or partially visible objects.
%This design neglects the unique conditions of indoor mobile robots, which, while navigating cluttered human--centric environments, repeatedly observe the same portions of the scene. Such repeated observations provide a natural proxy for consistency, which can be exploited to improve pseudo--label quality, reduce uncertainty, and enhance the recognition of small, occluded, or partially visible objects.

An early work is~\cite{continual_adaptation_2d3d}, which introduces the concept of multi--view consistency, leveraging the fact that model's predictions about the same locations of the environment must exhibit some level of similarity even if taken from multiple viewpoints. The method aggregates single--frame 2D semantic predictions of the robot into a 3D voxel map, which stores the probability for each voxel to belong to a certain semantic class. This dense representation of the environment is then used for rendering pseudo--labels that aggregate information from multiple viewpoints, thus removing the intrinsic noise of the frame--by--frame model's predictions. The approach is extended in~\cite{informative_planning}, where the authors propose to leverage the robot embodiment for data acquisition in specific environmental locations. The idea is to aggregate in the dense environmental map not only the semantic labels, but also the uncertainty estimation of the model's predictions. In this way, the robot can collect novel views focusing on areas with high uncertainty to improve the quality of the 3D annotations. In the recent work of~\cite{uda_neural_rendering}, the voxel map is substituted with a semantic NeRF, a more compact spatial representation of the environment that allows the rendering of novel views (RGB images) and their respective pseudo--labels to augment the training data.

%In this paper, we take a step forward in multi-view consistency, which, as shown by prior work, cannot resolve coherent misclassifications persisting across multiple frames. To address this limitation, we integrate a zero--shot, instance--aware refinement phase applied after 3D map rendering. This procedure enforces a single semantic class per object instance, also removing artifacts introduced during rendering.
While these methods present promising results, they still often struggle to resolve coherent misclassifications that persist across multiple frames, as we detail in Section \ref{sec:instance_labels}.
We advance multi--view consistency by addressing this limitation. We integrate a zero--shot, instance--aware refinement phase after 3D map rendering that enforces single semantic classes per object instance while removing rendering artifacts.

\begin{figure*}[!t]
\begin{center}
    \includegraphics[width=.9\linewidth]{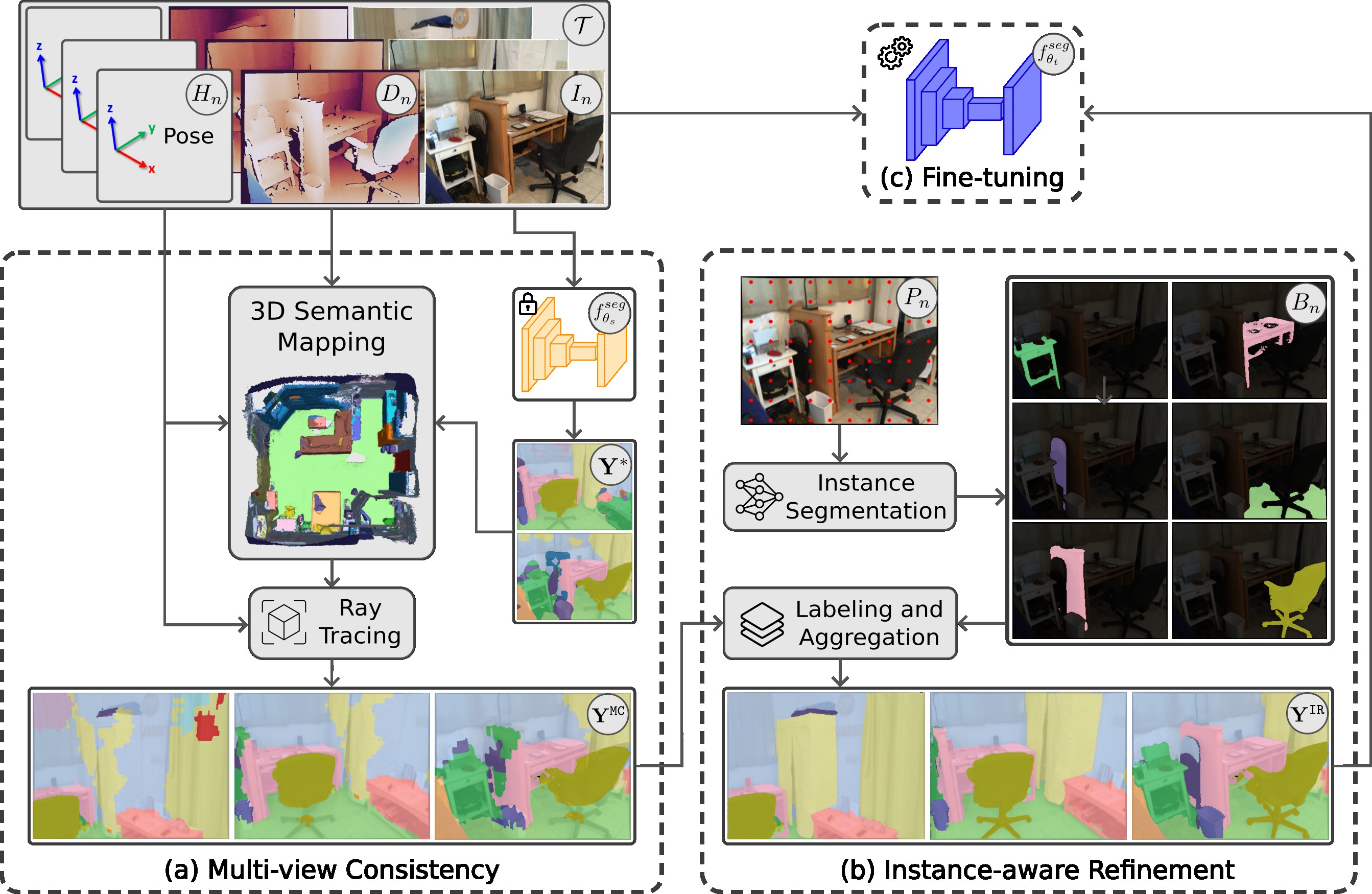}

    \caption{
    Our method (a) first aggregates the source model’s predictions $\mathbf{Y}^*$ into a 3D map to obtain multi--view consistent pseudo--labels $\mathbf{Y}^\texttt{MC}$. (b) These are then refined with a foundation model for instance segmentation. (c) The resulting pseudo--labels $\mathbf{Y}^\texttt{IR}$ are finally used to fine--tune the source model.
    }
    \label{fig:method_overview}
\end{center}
\vspace{-0.7cm}
\end{figure*}
\section{Method}

\subsection{Problem formulation}

%We consider a reference scenario where a mobile robot relies on a neural network to carry out semantic segmentation from images acquired with an on--board RGB--D camera. 

We consider a reference scenario in which a mobile robot relies on a neural network to perform semantic segmentation from images acquired with an on--board RGB camera, while depth information is obtained either from stereo or an RGB--D sensor. Formally, we describe the task with a function as $Y^* = f_{\theta_s}^{seg}(I)$ that predicts from an RGB image $I$ a semantic mask $Y^*$, where each pixel is assigned to an object category $o \in \mathcal{O}$. The network's parameters $\theta_s$ are pre--trained on a \emph{source} dataset $\mathcal{S}=\{\mathbf{I}^{s}, \mathbf{Y}^{s}\}$, composed of a set of images ($\mathbf{I}^s$) and their corresponding ground--truth semantic annotations ($\mathbf{Y}^s$) coming from multiple indoor scenes. We aim to mitigate the domain shift, that is the performance degradation that occurs when a robot's perception system encounters visual conditions, layouts, or object appearances that differ from its training data, experienced by the robot when deployed in a novel and previously unseen \emph{target} environment $t$, where no ground--truth semantic labels are available. From that environment, we suppose to have a sequence of $N$ perceptions $\mathcal{T} = \{\mathbf{I}^t, \mathbf{D}^t, \mathbf{H}^t\}$ acquired by the robot from multiple viewpoints, where each colored image $I_n \in \mathbf{I}^t$ and its corresponding depth $D_n \in \mathbf{D}^t$ are localized using the camera extrinsics $H_n \in \mathbf{H}^t$ (with $n \in \{1, \ldots, N\}$). Our goal is to adapt $f_{\theta_s}^{seg}$ for the target environment $t$ with no ground--truth and obtain a new semantic network $f_{\theta_t}^{seg}$ with improved performance in $t$. 

Fig.~\ref{fig:method_overview} presents a general overview of our proposed approach, which performs a self--supervised fine--tuning of $f_{\theta_s}^{seg}$ on data from the target scene $t$ using directly its predictions as pseudo--labels. Our method aims at improving the quality of these pseudo--labels to prevent the performance degradation typically caused by re--training with the raw model's outputs, which are particularly noisy due to domain shifts. Following~\cite{continual_adaptation_2d3d}, we start by aggregating the per--frame predictions of $f_{\theta_s}^{seg}$ in a 3D environmental representation, which is used for rendering multi--view consistent pseudo--labels (Section~\ref{sec:mvc_labels}, Fig.~\ref{fig:method_overview} (a)). Then, we add a novel instance refinement step to ensure instance--coherent semantic annotations and fix the artifacts produced by the rendering procedure (Section~\ref{sec:instance_labels}, Fig.~\ref{fig:method_overview} (b)). Finally, the refined pseudo--labels are used to fine--tune $f_{\theta_s}^{seg}$, obtaining $f_{\theta_t}^{seg}$ with improved performance in $t$ (Fig.~\ref{fig:method_overview} (c)).

\subsection{Multi--view consistent pseudo--labels}\label{sec:mvc_labels}
To retrieve multi--view consistent pseudo--labels we follow the method described in~\cite{continual_adaptation_2d3d}. From the sequence $\mathcal{T}$ we extract the set $\mathbf{Y}^* =\bigcup_{n=1}^{N} \{Y^*_n | Y^*_n = f_{\theta_s}^{seg}(I_n)\} $ containing the single--frame model's prediction for each image in $\mathbf{I}^t$. Then, we use Kimera Semantics~\cite{kimera} to aggregate $\mathcal{T}$ together with $\mathbf{Y}^*$ in a 3D semantically--annotated representation of the environment, in which the geometry is represented by a TSDF volume~\cite{voxblox} and the semantics are stored in an additional voxel layer. Each voxel maintains a distribution probability over the object categories, which is updated every frame $n$ by matching the 3D semantic points obtained by projecting the semantic pixels of $Y^*_n$ using the depth $D_n$ and the camera pose $H_n$.
%
%When the integration of all the measurements is complete, we export an high--resolution mesh, that is then ray--traced for each camera position $H_n$ to determine the first 3D intersection of each pixel in the camera plane with the mesh. This information is used to retrieve the class with the highest probability from the corresponding voxel. With this procedure, we produce the set $\mathbf{Y}^{\texttt{MC}}$ containing the multi--view consistent pseudo--labels rendered from each $H_n \in \mathbf{H}^t$.
%
After integrating all measurements, we export a high--resolution mesh and ray--trace it from each camera pose $H_n$ to find the first 3D intersection per pixel. The corresponding voxel provides the most likely class, yielding the set $\mathbf{Y}^\texttt{MC}$ of multi--view consistent pseudo--labels for all $H_n \in \mathbf{H}^t$
(see the example reported in Fig.~\ref{fig:method_overview}~(a)).

\subsection{Instance--aware refinement}\label{sec:instance_labels}
Despite the rendered pseudo--labels $\mathbf{Y}^{\texttt{MC}}$ show improved quality with respect to single--frame predictions (as they are persistent in consecutive viewpoints), they have some limitations that motivate our additional refinement step. %At first, the model's missclassifications, if repeated in multiple frames, are consolidated by the 3D aggregation procedure. 
This is a result of how robots perceive the environment while performing navigation. Robots often perceive objects only partially during navigation, leading to systematic errors in label aggregation. Take as a example a robot observing a cabinet from multiple views. When first observing it, the robot may only capture small corners or partially occluded views, which are difficult to recognize and thus are frequently misclassified as other categories, such as \texttt{wall}, as happens in the examples of Fig.~\ref{fig:method_overview}~(a). Because these partial views are encountered repeatedly, their incorrect labels end up dominating in the voxel--wise majority of the resulting 3D map. Later, once the cabinet is fully visible, the robot correctly classifies it, but those few correct predictions remain outnumbered. As a result, the final aggregated label is wrong, despite the robot having seen the cabinet clearly. Our instance--aware refinement step resolves this issue by leveraging SAM to group pixels into coherent object instances. Within each instance, the correct labels from full observations can outweigh the scattered errors from partial ones, allowing our method to propagate the right semantic category to the entire object, as shown in 
Fig.~\ref{fig:method_overview}~(b).
Additionally, the multi--view consistent pseudo--labels are affected by artifacts introduced by the 3D reconstruction (such as missing depth in reflective surfaces) and by the rendering process (e.g., inaccurate boundaries caused by the voxel discretization).

To address these limitations, we propose to further refine the pseudo--labels of $\mathbf{Y}^\texttt{MC}$ according to the object instances contained in their corresponding RGB images in $\mathbf{I}^t$.
To do this, our method leverages the zero--shot capabilities of Segment Anything (SAM)~\cite{sam, sam2}, a prompt--based foundation model for instance segmentation. 
From an RGB image $I_n$ and a list of $J$ prompts $P_n$ (in which each prompt $p_{n,j} \in P_n$ is expressed with a point and/or a bounding box), we use SAM to obtain a set of binary masks 
$$B_n = \bigcup_{j=1}^J \big\{b_{n,j}|b_{n,j} = f^{sam}_\theta(I_n, p_{n, j}) \big\},$$
    where $f^{sam}_\theta$ denotes the SAM model with pre--trained parameters $\theta$. Inside each $b_{n, j}$, a pixel $u$ assumes the value $b_{n, j}(u) = 1$ if it lies inside the instance identified by SAM from the relative prompt $p_{n,j}$, otherwise $b_{n, j}(u) = 0$.

For prompting SAM, we propose two alternative strategies, which we later compare. The first one, named \texttt{grid}, uses a fixed prompt, which is created once and used for every frame. In this strategy, $P_n$ is defined as a list of points uniformly arranged in a grid structure on the image plane, where $d$ specifies the distance between adjacent points in the horizontal and vertical axes. In the second one, called \texttt{informed}, the prompt is procedurally extracted for each frame using its rendered pseudo--label. At first, a pseudo--label $Y^\texttt{MC}_n$ is partitioned into clusters, where each cluster is a connected region of pixels sharing the same semantic class. Then, for each cluster larger than a percentage $a$ of the image area, it calculates the bounding box and its centroid. $P_n$ is finally composed by aggregating the bounding box/centroid pairs derived from all clusters.

Independently from the prompting strategy, our method proceeds by combining the object instances of $B_n$ with the classes contained in the corresponding pseudo--label $Y_n^\texttt{MC}$. To do this, for each instance mask $b_{n,j} \in B_n$, we extract the most frequent object category according $Y_n^\texttt{MC}$, formally defined as 
$$o^*_{n, j} = \arg\max_{o\in \mathcal{O}} \sum_{u\in \Omega} b_{n, j}(u)\mathds{1}_{\{Y_n^{\texttt{MC}}(u) = o\}},$$
where $\mathds{1}_{\{\cdot\}}$ is the indicator function and $\Omega$ are the pixels indices.
All the binary masks $b_{n,j} \in B_n$ are then merged, together with their relative object category $o^*_{n,j}$, in a single instance--aware pseudo--label $Y^{\texttt{IR}}_n$. At the beginning, we set $Y^{\texttt{IR}}_n = Y^{\texttt{MC}}_n $. Then, for each mask $b_{n,j} \in B_n$, we update $Y^{\texttt{IR}}_n$ by overriding the pixels's category inside the region defined by $b_{n,j}$ with $o^*_{n,j}$, formally

$$Y^{\texttt{IR}}_n(u) = \begin{cases}
    o^*_{n,j} & \text{if } b_{n, j}(u) =1\\
    Y^{\texttt{IR}}_n(u) & \text{if } b_{n, j}(u) =0

\end{cases}\quad\forall u\in \Omega.$$

Note that this step is sequentially applied for all  $j \in \{1, \ldots, J\}$, meaning that previously added semantic labels can, in principle, be replaced in later stages. When $B_n$ is derived using the \texttt{informed} prompting strategy, this replacement is not possible: the object instances are disjoint by construction since they are extracted from non--overlapping clusters of labels. In contrast, when the \texttt{grid} strategy is used, the order in which the instances $b_{n,j}$ are aggregated might impact on the quality of the pseudo--labels. This is because $B_n$ contains a high number of overlapping instances, meaning that small objects may be completely overwritten by larger ones aggregated later. To avoid this, we sort the $B_n$ in descending order of instance dimension, thus preserving the fine--grained details identified by SAM when prompted with the \texttt{grid} strategy.
This entire instance--oriented refinement procedure is performed for all the images/pseudo--labels pairs in $\mathbf{I}^t$ and $\mathbf{Y}^{\texttt{MC}}$. The resulting instance--aware annotations (for all the measurements in $\mathcal{T}$) are contained in the set $\mathbf{Y}^\texttt{IR}$.
Finally, we use refined pseudo--labels $\mathbf{Y}^\texttt{IR}$ to finetune the model, thus doing UDA, as shown in Fig.~\ref{fig:method_overview} (c).

\section{Evaluation}

\subsection{Experimental setting}

\noindent\textbf{Dataset} For the evaluation of our proposed method, we rely on ScanNet~\cite{scannet}, a real--world dataset containing $1513$ sequences acquired in $707$ different indoor environments. Each sequence includes RGB--D image pairs %(acquired by a calibrated hardware setup composed of a Structure sensor~\cite{structure_sensor} and an iPad) 
with their relative camera poses, estimated using BundleFusion~\cite{bundle_fusion}. The dataset also provides the per--frame ground--truths, which are manually annotated with the NYU40 object categories~\cite{nyu40}. In all the experiments, we rescale the images to a resolution of $320\times240$ pixels, to be compliant with the hardware setup of a low--powered mobile robot.  

\noindent\textbf{Network pre--training.} Similarly to~\cite{continual_adaptation_2d3d, uda_neural_rendering}, the scenes from $11$ to $707$ represent our source domain and are used to initialize the weights of the neural network for semantic segmentation $f_{\theta_s}^{seg}$. For simplicity, we rely on the DeepLabV3~\cite{deeplabv3} (with a ResNet--101~\cite{resnet} as backbone) provided by~\cite{uda_neural_rendering}, which was pre--trained on a dataset $\mathcal{S}$ counting $\approx$ $25\text{k}$ images/ground--truths pairs from environments $11$--$707$, sampling one frame every $100$. $\mathcal{S}$ was randomly split in $\approx$ $20\text{k}$ and $\approx$ $5\text{k}$ samples for training and validation, respectively. The network is optimized using Adam for $150$ epochs, with a batch size of $4$. To measure the quality of the predictions, we use the Intersection over Union averaged over the objects categories (mIoU).

\noindent\textbf{Network adaptation.} Scenes $1$--$10$, that represent the target environments, are used to obtain $f_{\theta_t}^{seg}$ by fine--tuning $f_{\theta_s}^{seg}$ with the pseudo--labels provided by our method.  We consider the first sequence for each environment, from which the first $80\%$ of the frames are used to produce the pseudo--labels and fine--tuning $f_{\theta_s}^{seg}$, while the last $20\%$ is used for testing.  This is the same method adopted in~\cite{continual_adaptation_2d3d} and~\cite{uda_neural_rendering}, which we replicate but also extend to achieve a more representative evaluation. Although training and testing frames are temporally disjoint, they are sampled from the same sequence within a given environment. In contrast, real--world deployments typically involve two distinct runs: one to collect data for unsupervised adaptation and another to evaluate performance in a different trajectory. To better reflect this setting, we perform an additional evaluation in which two separate sequences are used for each environment: one for pseudo--label generation and fine--tuning, and the other for testing. (We cannot run this protocol on environments $5$, $8$, and $9$, as only a single sequence is available in the dataset.) Overall, this experimental design aims at better reflecting the practical conditions of robotic operation: a model $f_{\theta_s}^{seg}$ is first pre--trained on a large source dataset, the robot then collects an initial set of observations (e.g., while mapping the environment) to generate pseudo--labels and adapt $f_{\theta_s}^{seg}$ in an unsupervised fashion. Finally, the adapted model $f_{\theta_t}^{seg}$ is tested on new trajectories.

%We envision real--world robotic deployments where one run might be used to collect data for unsupervised adaptation and then a completely different run is used to perform testing.

%this approach does not completely reflect the real--world deployment of a robot as the evaluation is performed considering a scene's subportion. 

%To overcome this, we perform an additional evaluation using two different sequences for each environment: the first one is used to produce pseudo--labels and fine--tuning while the second one is reserved for testing. (We cannot perform this additional evaluation in environments $5$, $8$, and $9$ as they have only a single sequence in the dataset.)
%We remark that our experimental campaign is targeted to replicate the conditions experienced by a robot in a real--world deployment: at first, $f_{\theta_s}^{seg}$ is initialized using a large source dataset, then the robot performs a data collection in the first phase of its setup (e.g., while mapping the environment) that is used to produce the pseudo--labels and fine--tune $f_{\theta_s}^{seg}$ in an unsupervised fashion, and finally the performance of $f_{\theta_t}^{seg}$ is tested in a different trajectory. 

\noindent\textbf{Instance--aware refinement.} To generate the multi--view consistent pseudo--labels $\mathbf{Y}^{\texttt{MC}}$, we leveraged the open--source implementation of~\cite{continual_adaptation_2d3d}. We set the voxel size of Kimera Semantics~\cite{kimera} to $3\text{cm}$ and $5\text{cm}$, to evaluate the performance under different resolutions, following the works of \cite{continual_adaptation_2d3d, uda_neural_rendering}. More precisely, setting the voxel size to $3\text{cm}$ ($5\text{cm}$) creates a more (less) detailed 3D representation, requiring higher (lower) processing power for real--time robot deployment. 
%The code for generating the 3D semantic map with Kimera is written in ROS.%, meaning that it can be run online by a mobile robot by changing the input measurements. 
The object instances $B_n$ are obtained using the Ultralytics' implementation of Segment Anything v2~\cite{sam2}. For prompting SAM, the points of the \texttt{grid} strategy are generated with a distance $d=32$, while the cluster of labels smaller than $a = 0.1\%$ of the image size are not considered in the \texttt{informed} approach. 
To evaluate the performance improvements related to the two prompting strategies, we compare three versions of $f_{\theta_t}^{seg}$ for each environment: (i) using only \texttt{grid}, (ii) using only \texttt{informed}, and (iii) combining the two. For the single--method settings, $f_{\theta_s}^{seg}$ is optimized using $\mathbf{Y}^\texttt{IR}$ and their corresponding RGB images for $10$ epochs using Adam with a learning rate of $10^{-5}$, and batch size of $4$. When the methods are combined, the model processes the same RGB image twice inside the same epoch, each time with different pseudo--labels. To maintain a constant training budget, epochs are halved from $10$ to $5$. Random flipping and color jitter are used as data augmentation.

\noindent\textbf{Baseline.} The baseline we use to evaluate our approach is the multi--view consistency~\cite{continual_adaptation_2d3d}, that is also an important building block of our pipeline, as detailed in Section~\ref{sec:mvc_labels}. For a fair comparison, we replicate the results by re--generating the pseudo--labels from scratch using the aforementioned pre--trained DeepLabV3. With the multi--view consistent annotations $\mathbf{Y}^{\texttt{MC}}$ and their relative RGB images, we fine--tune a model $f_{\theta_t}^{seg}$ for each target environment, using the Adam optimizer with a learning rate of $10^{-5}$ and batch size of $4$ for $10$ epochs.

\vspace{-0.2cm}

\subsection{Pseudo--labels improvement}
\label{sec:pseudo_labels_improvement}
\setlength{\tabcolsep}{0.65em}
\begin{table}[]
\caption{Pseudo--labels performance}
\vspace{-0.2cm}
\scriptsize
\centering

\begin{tabular}{c|c|cc|cccc}
\toprule
&&\multicolumn{2}{c|}{Baseline (5cm)} & \multicolumn{4}{c}{Our method}\\
Env &$\mathbf{Y}^*$ & $\mathbf{Y}^\texttt{MC}$& $\Delta_{\mathbf{Y}^*}$   &$\mathbf{Y}^\texttt{IR}_\texttt{G}$ &$\Delta_{\mathbf{Y}^\texttt{MC}}$ & $\mathbf{Y}^\texttt{IR}_\texttt{I}$ & $\Delta_{\mathbf{Y}^\texttt{MC}}$\\
\midrule
1 &	41,2&	48,1&	16,7\%&	   	\bestvalue{52,3}&	8,7\% &  \secondvalue{49,1}&	2,1\%\\
2&	\bestvalue{34,8}&	28,8&	-17,2\%&		\secondvalue{32,5}&	12,8\% & 31,5&	9,4\%\\
3&	23,7&	\secondvalue{26}&	     9,7\%&	    	25,4&	-2,3\% & \bestvalue{26,5}&	1,9\%\\
4&	62,8&	63,9&	1,8\%&	    	\bestvalue{65,2}&	2,0\%& \secondvalue{64,7}&	1,3\%\\
5&	\bestvalue{49,8}&	42,6&	-14,5\%&		\secondvalue{45,5}&	6,8\% & 44,9&	5,4\%\\
 6&	48,7&	49,3&	1,2\%&	    	\secondvalue{52,6}&	6,7\% &\bestvalue{53,3}&	8,1\%\\
 7&	40,3&	48,4&	20,1\%&	    	\secondvalue{48,8}&	0,8\% & \bestvalue{50,8}&	5,0\%\\
 8&	31,4&	34,8&	10,8\%&	   	\secondvalue{35,4}&	1,7\% &\bestvalue{36,4}&	4,6\%\\
 9&	31,8&	\bestvalue{32,8}&	3,1\%&	   	30,8&	-6,1\% & \secondvalue{32,1}&	-2,1\%\\
 10&	52,1&	55,8&	7,1\%&	   	\bestvalue{59,6}&	6,8\% & \secondvalue{58,8}&	5,4\%\\\midrule
Avg	&41,7&	43,1&	3,9\%&		\secondvalue{44,8}&	3,8\% & \bestvalue{44,8}&	4,1\%\\

\bottomrule
\multicolumn{8}{l}{}\\[-0.2cm]
\multicolumn{8}{p{0.43\textwidth}}{mIoU (\bestvalue{best} and \secondvalue{second best} values) of the pseudo--labels from our method ($\mathbf{Y}^\texttt{IR}_\texttt{G}$ and $\mathbf{Y}^\texttt{IR}_\texttt{I}$) against the baseline ($\mathbf{Y}^\texttt{MC}$) and the pre--trained model ($\mathbf{Y}^*$).}
\end{tabular}

%\caption{mIoU of the instance--aware pseudo--labels with our method prompted with \texttt{grid} ($\mathbf{Y}^\texttt{IR}_\texttt{G}$) and \texttt{informed} ($\mathbf{Y}^\texttt{IR}_\texttt{I}$) against the baseline ($\mathbf{Y}^\texttt{MC}$) and the predictions of the pre--trained model ($\mathbf{Y}^*$). Best and second best performances are respectively in \bestvalue{bold} and \secondvalue{underlined}. }
%\caption{mIoU of the instance--aware pseudo--labels with our method prompted with \texttt{grid} ($\mathbf{Y}^\texttt{IR}_\texttt{G}$) and \texttt{informed} ($\mathbf{Y}^\texttt{IR}_\texttt{I}$) against the baseline ($\mathbf{Y}^\texttt{MC}$) and the predictions of the pre--trained model ($\mathbf{Y}^*$). Best and second best performances are respectively in \bestvalue{bold} and \secondvalue{underlined}. }
\label{tab:psuedo_labels_quality}
\vspace{-0.6cm}

\end{table}

This section evaluates the quality of the instance--aware pseudo--labels produced by our method compared to those of the baseline. The quantitative results are reported in Table~\ref{tab:psuedo_labels_quality}, which shows the precision of the pseudo--labels (measured in mIoU) obtained from the first 80\% of the first sequence for each environment, using a voxel size of $5\text{cm}$. We refer to the refined pseudo--labels as $\mathbf{Y}_\texttt{G}^\texttt{IR}$ or $\mathbf{Y}_\texttt{I}^\texttt{IR}$ if obtained using the \texttt{grid} or the \texttt{informed} prompting strategy.

\setlength{\tabcolsep}{0.43em}

\begin{table*}[t]
\caption{Performance of the adapted model when training -- testing use the 80\% -- 20\% of the same sequence of each environment}
\vspace{-0.2cm}
\scriptsize
\centering

\begin{tabular}{c|c|cc|cccccc||c|cc|cccccc}
\toprule
&&\multicolumn{2}{c|}{Baseline (5cm)} & \multicolumn{6}{c||}{Our method}&&\multicolumn{2}{c|}{Baseline (3cm)} & \multicolumn{6}{c}{Our method}\\
Env & \PT & \MC& $\Delta_{\text{\PT}}$ &\IRG & $\Delta_{\text{\MC}} $& \IRI & $\Delta_{\text{\MC}} $ & \IRGI & $\Delta_{\text{\MC}} $& \PT& \MC& $\Delta_{\text{\PT}}$ &\IRG & $\Delta_{\text{\MC}} $& \IRI & $\Delta_{\text{\MC}} $ & \IRGI & $\Delta_{\text{\MC}} $ \\
\midrule
 
 1	&    44,2&	48,2&	9,0\%&	    \bestvalue{51}&	    5,8\%&	   49,3&	  2,3\%& \secondvalue{50,9} & 5,6\% &	  44,2&	47,6&	7,7\%&	    50,1&	5,3\%&	    \secondvalue{50,4}&	5,9\% &\bestvalue{51}	&7,1\%\\[0.05cm]
 2	 &   \bestvalue{36,2}&	31,3&	-13,5\%&	33,6&	7,3\%&	   34,4&	  9,9\%&	\secondvalue{34,8} &	11,2\% &    \bestvalue{36,2}&	34,4&	-5,0\%&	    34,5&	0,3\%&	    \secondvalue{34,9}&	1,5\% &33,8&	-1,7\%\\[0.05cm]
 3	&    22,9&	21&	    -8,3\%&	    21,4&	1,9\%&	   \bestvalue{23,2}&	  10,5\%&	\secondvalue{22,9}&	9,0\%  &   22,9&	21,7&	-5,2\%&	    22,6&	4,1\%&	    \secondvalue{23,1}&	6,5\%&\bestvalue{23,2}	&6,9\%\\[0.05cm]
 4&	 50,1&	51,1&	2,0\%&	    52,5&	2,7\%&	   \secondvalue{52,5}&	  2,7\%&	\bestvalue{53,4}&	4,5\% &    50,1&	52,4&	4,6\%&	    \secondvalue{52,9}&	1,0\%&	    49,5&	-5,5\%&\bestvalue{53,7}	&2,5\%\\[0.05cm]
 5&	 39,7&	40,1&	1,0\%&	    \secondvalue{44,5}&	11,0\%&	   44,4&      10,7\%&	 \bestvalue{44,7}&	11,5\%  &   39,7&	45&	    13,4\%&	    \secondvalue{47,2}&	4,9\%&	    46,1&	2,4\%&\bestvalue{47,3}&	5,1\%\\[0.05cm]
 6&	 \bestvalue{35}&    31,9&	-8,9\%&	    33,9&	6,3\%&	   34,1&      6,9\%&	\secondvalue{34,9}&	9,4\% &    35&	    33,5&	-4,3\%&	    \secondvalue{39,5}&	17,9\%&	    38,8&	15,8\%&\bestvalue{40,1}&	19,7\%\\[0.05cm]
 7&	 56,7&	\secondvalue{63,5}&	12,0\%&	    60,7&	-4,4\%&	   \bestvalue{63,5}&      0,0\%&	60,9&	-4,1\% &    56,7&	\bestvalue{63}&	    11,1\%&	    \secondvalue{60,9}&	-3,3\%&	    56,7&	-10,0\%&59,5&	-5,6\%\\[0.05cm]
 8&	  \bestvalue{29,5}&	25,1&	-14,9\%&	25,2&	0,4\%&	   26,3&      4,8\%&	\secondvalue{27,6} &	10,0\%  &    \bestvalue{29,5}&	25,1&	-14,9\%&	24,1&	-4,0\%&	    \secondvalue{27}&	   7,6\%&25,7&	2,4\%\\[0.05cm]
 9	&    55,9&	65,3&	16,8\%&	    \bestvalue{70,2}&	7,5\%&     69&  	  5,7\%&	\secondvalue{69,2}&	6,0\% &    55,9&	66,9&	19,7\%&	    \bestvalue{70,9}&	6,0\%&	    69,8&	4,3\%&\secondvalue{69,9}	&4,5\%\\[0.05cm]
 10&	73,4&	71,9&	-2,0\%&	    \bestvalue{75,3}&	4,7\%&     74,7&      3,9\%&	 \secondvalue{75}&	4,3\% &   73,4&	73,2&	-0,3\%&	    \bestvalue{75,7}&	3,4\%&	    \secondvalue{75,6}&	3,3\%&75,3&	2,9\%\\\midrule
Avg&	     44,4&	44,9&	-0,7\%&	    46,8&	4,3\%&	 \secondvalue{46,8}&      5,7\%&	\bestvalue{47,4}&	6,7\% &    44,4&	46,3&	2,7\%&	    \secondvalue{47,8}&	3,6\%&	    47,2&	3,2\%&\bestvalue{48,0}&	4,4\%\\
\bottomrule
\multicolumn{19}{l}{}\\[-0.2cm]
\multicolumn{19}{p{0.9\textwidth}}{mIoU (\bestvalue{best} and \secondvalue{second best} values) of $f_{\theta_t}^{seg}$ trained with our method (\IRG, \IRI, and \IRGI) against the baseline (\MC), and the pre--trained model (\PT).}
\end{tabular}

%\caption{Improvements of $f_{\theta_t}^{seg}$ trained using the instance--aware pseudo--labels using \texttt{grid} (\IRG), \texttt{informed} (\IRI), and the combination (\IRGI) of the strategies compared with the fine--tuning using the multi--view consistent annotations (\MC) and the outputs of the pre--trained model $f_{\theta_s}^{seg}$ (\PT). The voxel size is set to 5cm and 3cm. Training and testing is performed using the first 80\% and the last 20\% of the first sequence of each environment. Best and second best performances are in \bestvalue{bold} and \secondvalue{underlined}.}
\label{tab:performance_20}
\vspace{-0.5cm}

\end{table*}

The metrics show that our instance--refinement step improves the multi--view consistent annotations $\mathbf{Y}^\texttt{MC}$ of the baseline, on average by $3.8\%$ and $4.1\%$ for \texttt{grid} and \texttt{informed}, respectively. Despite the baseline improves the raw model's predictions $\mathbf{Y}^*$ by $3,9\%$ on average, its impact is not stable and changes according to the environment. In some cases, the multi--view consistency yields substantial improvements (such as in environments $1$ and $7$ with an increment of $16,7\%$ and $20.1\%$), while in others it strongly degrades the quality of the per--frame predictions (e.g., in environments $2$ and $5$ with a decrease of $-17,2\%$ and $-14,5\%$). In contrast, our instance--aware refinement step increases the quality of $\mathbf{Y}^\texttt{MC}$ in all environments, with only marginal decreases observed in scenes $3$ and $9$. As shown in Fig.~\ref{fig:pseudo_labels_5cm}, our approach substantially improves the precision of the pseudo--labels by propagating the object classes according to instances identified in the RGB perception, and fixes the rendering artifacts produced by the ray tracing.

\renewcommand{\imageone}{5cm/grid/scene0005_00/64.png}
\renewcommand{\imagetwo}{5cm/grid/scene0004_00/570.png}
\renewcommand{\imagethree}{5cm/informed/scene0002_00/443.png}

\renewcommand{\imagefour}{5cm/informed/scene0003_00/632.png}

\renewcommand{\ww}{0.22\linewidth}

\begin{figure}[]
\begin{tabular}{@{}p{0.25cm}c@{ }c@{ }c@{ }c@{ }}

& \multicolumn{2}{c}{{\small\texttt{grid}}} & \multicolumn{2}{c}{{\small\texttt{informed}}}\\
{\small$\mathbf{I}^t$}&
\raisebox{-.4\height}{\includegraphics[trim=0 480 640 0, clip, width=\ww]{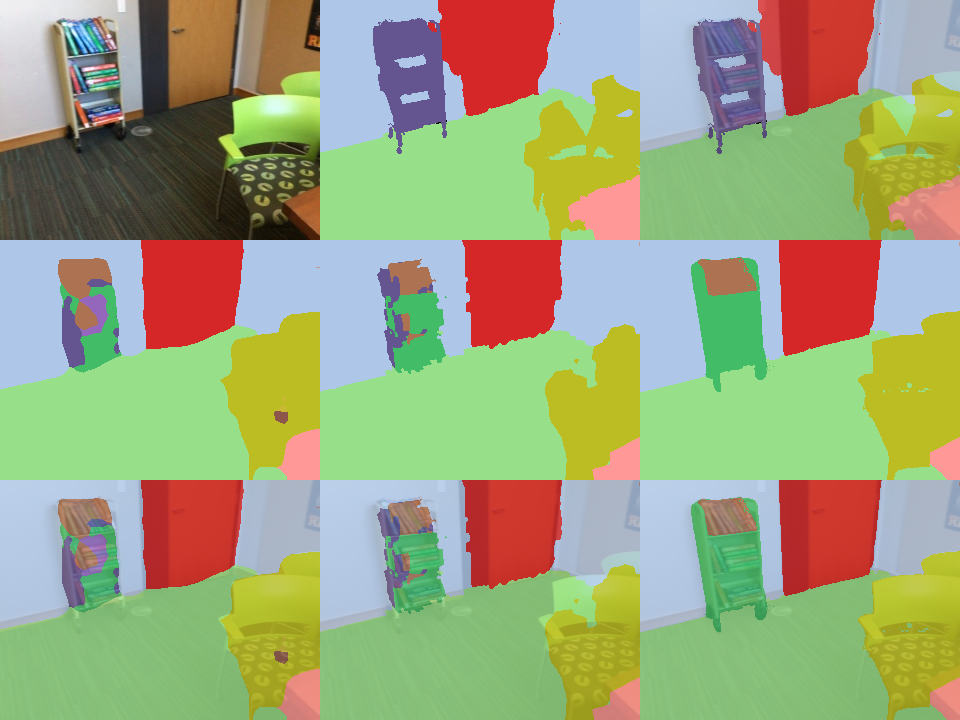}}&
\raisebox{-.4\height}{\includegraphics[trim=0 480 640 0, clip, width=\ww]{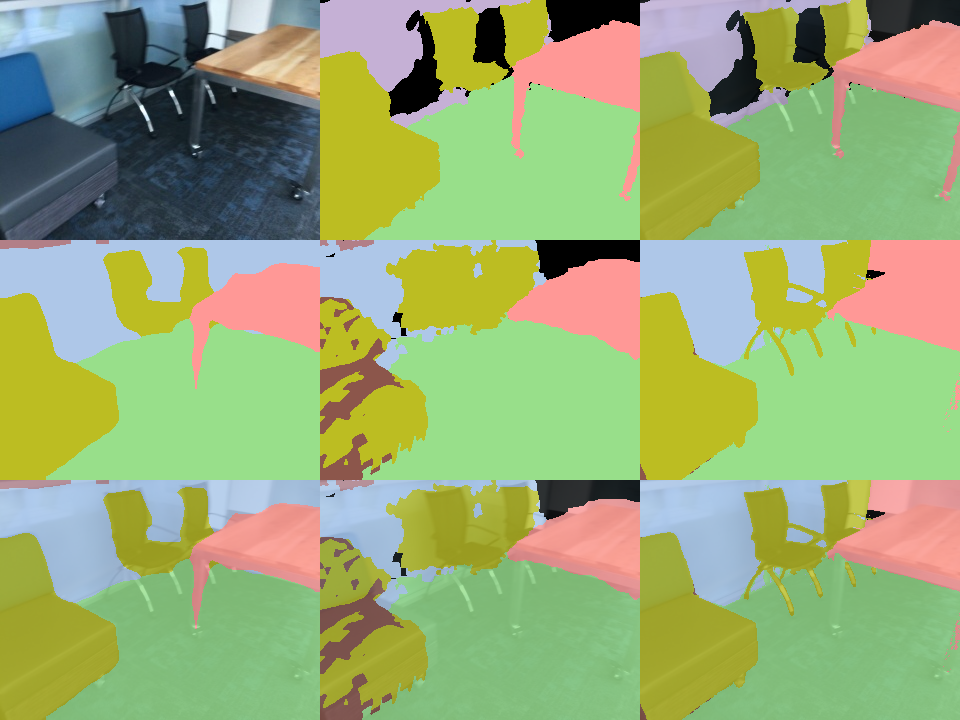}}&
\raisebox{-.4\height}{\includegraphics[trim=0 480 640 0, clip, width=\ww]{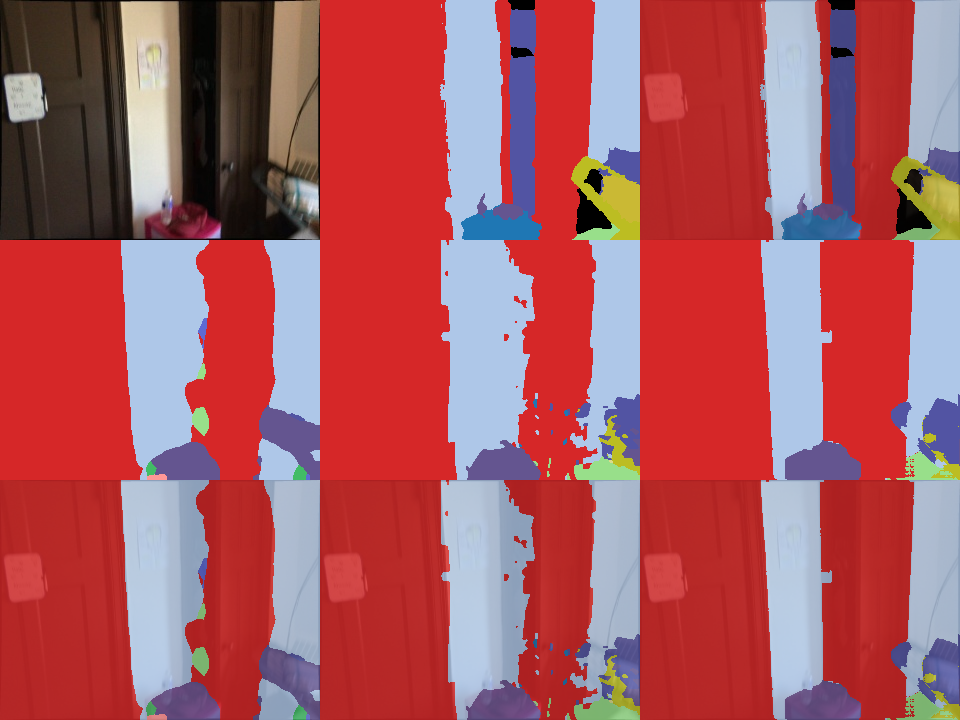}}&
\raisebox{-.4\height}{\includegraphics[trim=0 480 640 0, clip, width=\ww]{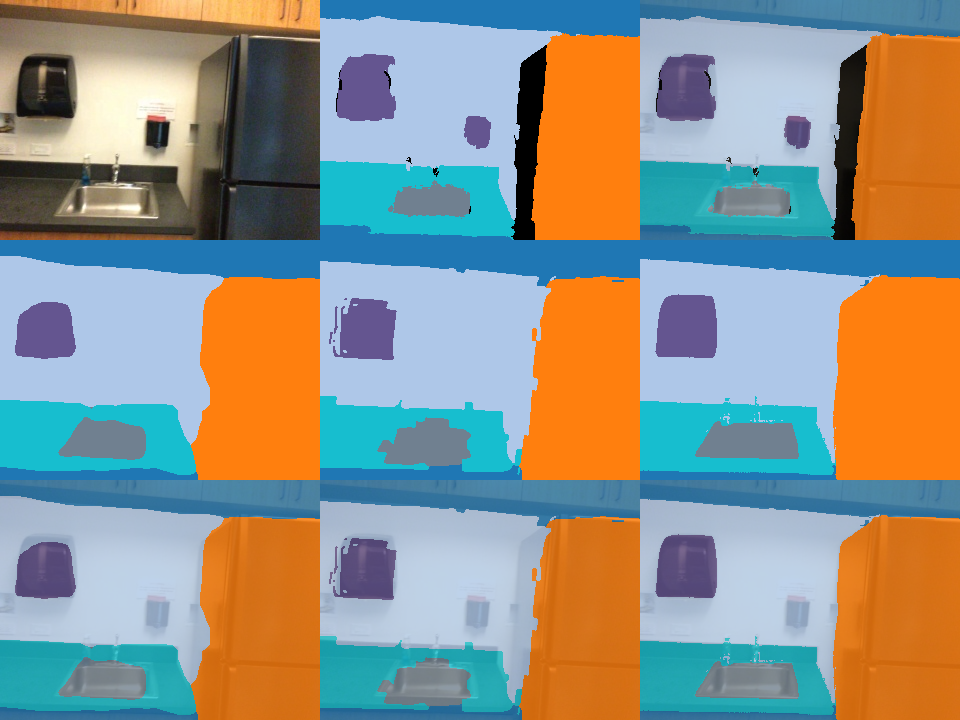}}\\[0.55cm]

{\small$\mathbf{Y}^*$}&
\raisebox{-.4\height}{\includegraphics[trim=0 0 640 480, clip,width=\ww]{contents/images/pseudo_labels/\imageone}}&
\raisebox{-.4\height}{\includegraphics[trim=0 0 640 480, clip,width=\ww]{contents/images/pseudo_labels/\imagetwo}}&
\raisebox{-.4\height}{\includegraphics[trim=0 0 640 480, clip,width=\ww]{contents/images/pseudo_labels/\imagethree}}&
\raisebox{-.4\height}{\includegraphics[trim=0 0 640 480, clip,width=\ww]{contents/images/pseudo_labels/\imagefour}}\\[0.55cm]

{\small$\mathbf{Y}^\texttt{MC}$}&
\raisebox{-.4\height}{\includegraphics[trim=320 0 320 480, clip,width=\ww]{contents/images/pseudo_labels/\imageone}}&
\raisebox{-.4\height}{\includegraphics[trim=320 0 320 480, clip,width=\ww]{contents/images/pseudo_labels/\imagetwo}}&
\raisebox{-.4\height}{\includegraphics[trim=320 0 320 480, clip,width=\ww]{contents/images/pseudo_labels/\imagethree}}&
\raisebox{-.4\height}{\includegraphics[trim=320 0 320 480, clip,width=\ww]{contents/images/pseudo_labels/\imagefour}}\\[0.55cm]
    
{\small$\mathbf{Y}^\texttt{IR}$}&

\raisebox{-.4\height}{\includegraphics[trim=640 0 0 480, clip,width=\ww]{contents/images/pseudo_labels/\imageone}}&
\raisebox{-.4\height}{\includegraphics[trim=640 0 0 480, clip,width=\ww]{contents/images/pseudo_labels/\imagetwo}}&
\raisebox{-.4\height}{\includegraphics[trim=640 0 0 480, clip,width=\ww]{contents/images/pseudo_labels/\imagethree}}&
\raisebox{-.4\height}{\includegraphics[trim=640 0 0 480, clip,width=\ww]{contents/images/pseudo_labels/\imagefour}}\\[0.55cm]

{\small$\mathbf{Y}^\texttt{GT}$}&

\raisebox{-.4\height}{\includegraphics[trim=640 480 0 0, clip,width=\ww]{contents/images/pseudo_labels/\imageone}}&
\raisebox{-.4\height}{\includegraphics[trim=640 480 0 0, clip,width=\ww]{contents/images/pseudo_labels/\imagetwo}}&
\raisebox{-.4\height}{\includegraphics[trim=640 480 0 0, clip,width=\ww]{contents/images/pseudo_labels/\imagethree}}&
\raisebox{-.4\height}{\includegraphics[trim=640 480 0 0, clip,width=\ww]{contents/images/pseudo_labels/\imagefour}}\\

\end{tabular}
    %\caption{Examples of the instance--aware pseudo--labels generated by our method ($\mathbf{Y}^\texttt{IR}$) obtained with the \texttt{grid} (left) and \texttt{informed} (right) strategies compared with the multi--view consistent ($\mathbf{Y}^\texttt{MC}$) and the raw ($\mathbf{Y}^*$) annotations. $\mathbf{Y}^t$ and $\mathbf{Y}^\texttt{GT}$ are, respectively, the RGB images and their relative ground--truth segmentations.}
    \caption{Instance--aware pseudo--labels ($\mathbf{Y}^\texttt{IR}$) generated with \texttt{grid} (left) and \texttt{informed} (right), compared to multi--view consistent ($\mathbf{Y}^\texttt{MC}$) and raw ($\mathbf{Y}^*$) ones. $\mathbf{I}^t$ and $\mathbf{Y}^\texttt{GT}$ denote the RGB images and ground truths.}
    \label{fig:pseudo_labels_5cm}
    \vspace{-0.7cm}
\end{figure}

Another interesting outcome from the results of Table~\ref{tab:psuedo_labels_quality} is that, although the two prompting strategies achieve the same average precision, their performances vary across environments. This can be explained by their markedly different behaviors, which are, to some extent, complementary. 

The \texttt{grid} approach exploits all the foundational model’s strengths in instance segmentation by using a fixed grid of points for each frame. This setting makes this method particularly effective in challenging situations where the objects have a complex shape or the multi--view consistent pseudo--labels are particularly noisy. Examples are reported in Fig.~\ref{fig:promts_diff}, respectively in the first and second columns, where we can see that the \texttt{grid} approach precisely identifies the instances of a guitar and a chair, which is partially occluded by a pillow. Compared to the automatic approach, the \texttt{informed} prompts derived from $\mathbf{Y}^\texttt{MC}$ produce less precise object instances. In the first case, the bounding box cuts the guitar’s neck, while in the second one, the couch and the chair are grouped in the same prompt (because they form a single cluster), but only the couch is segmented. %\textcolor{red}{Further results are available on the website.}

\renewcommand{\imageone}{1020}
\renewcommand{\imagetwo}{0}
\renewcommand{\imagethree}{690}
\renewcommand{\imagefour}{409}

\renewcommand{\ww}{0.21\linewidth}

\begin{figure}[]
\begin{tabular}{@{}p{0.25cm}c@{ }c@{ }c@{ }c@{ }}

{\small$\mathbf{I}^t$}&
\raisebox{-.4\height}{\includegraphics[width=\ww]{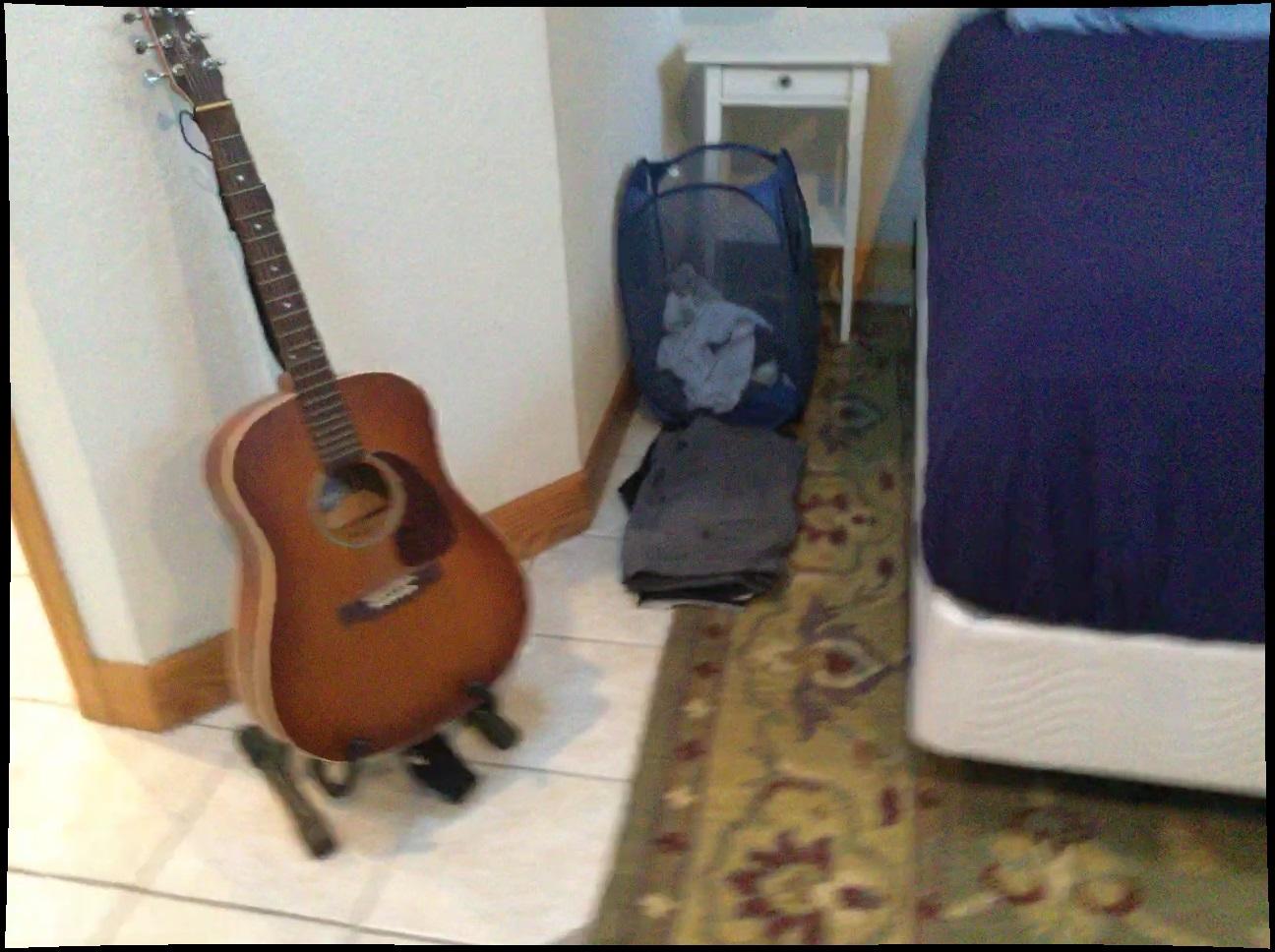}}&
\raisebox{-.4\height}{\includegraphics[width=\ww]{contents/images/prompts_differences/\imagetwo/color.jpg}}&
\raisebox{-.4\height}{\includegraphics[width=\ww]{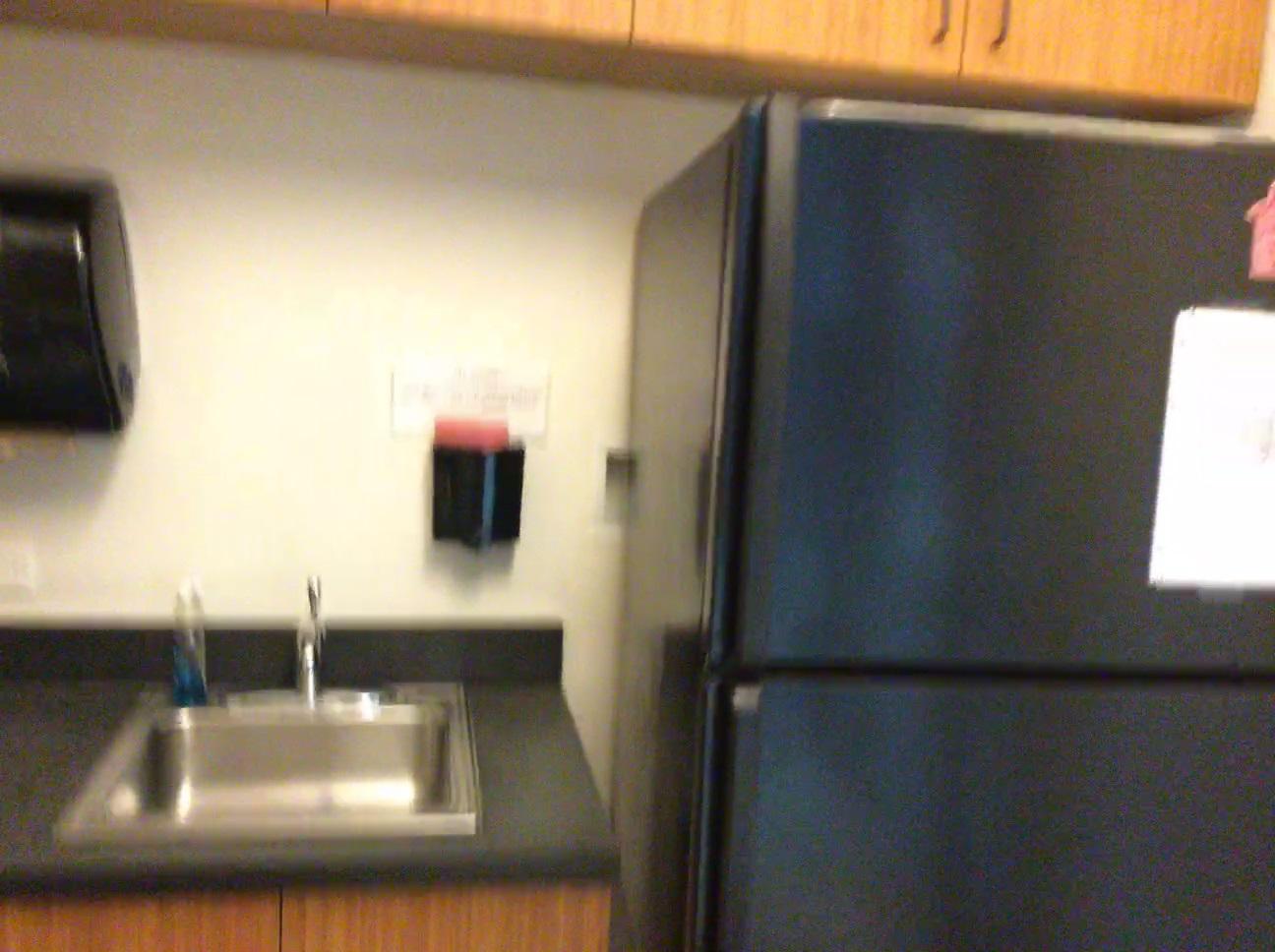}}&
\raisebox{-.4\height}{\includegraphics[width=\ww]{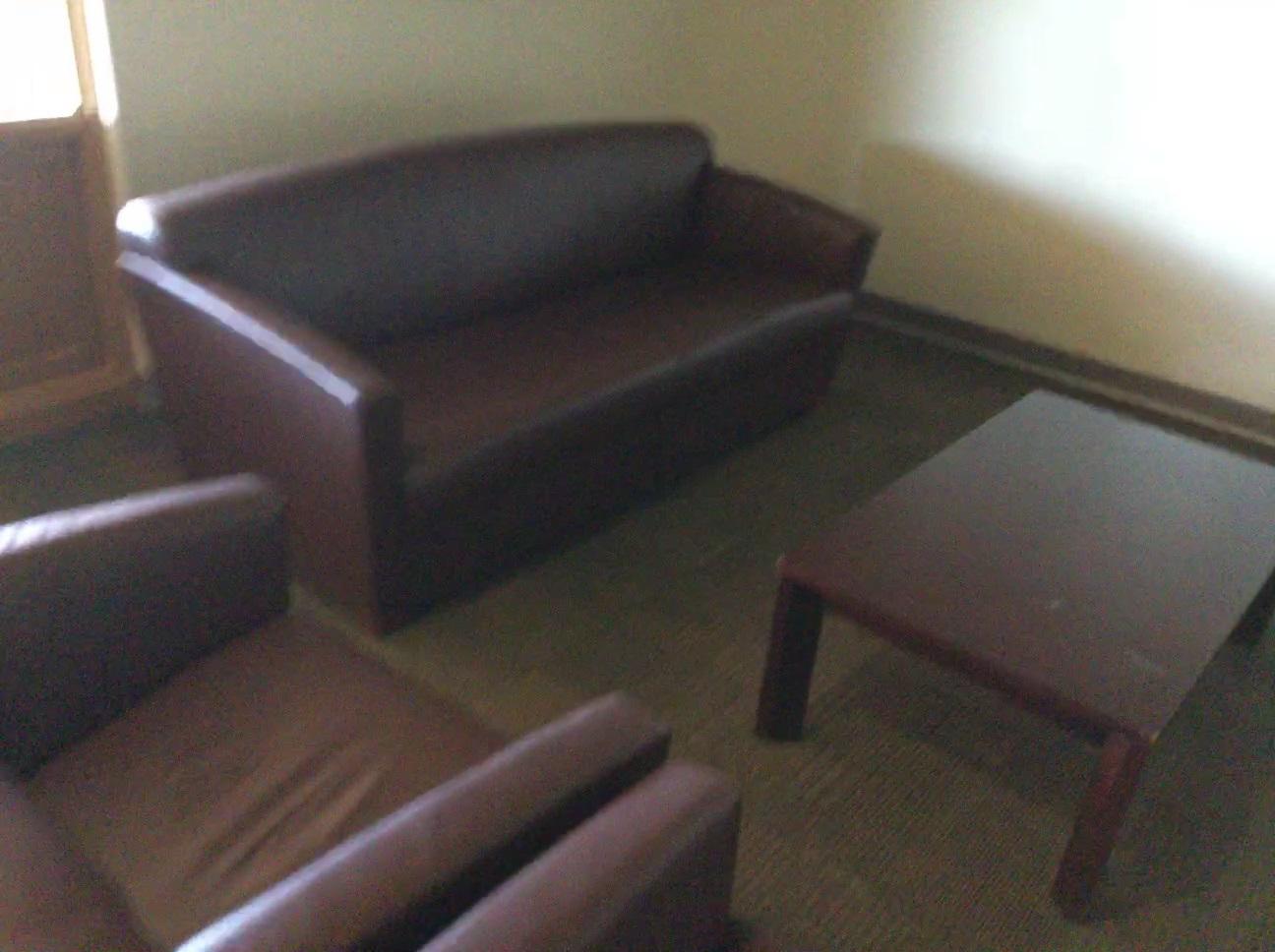}}\\[0.55cm]

{\small$\mathbf{Y}^\texttt{MC}$}&
\raisebox{-.4\height}{\includegraphics[width=\ww]{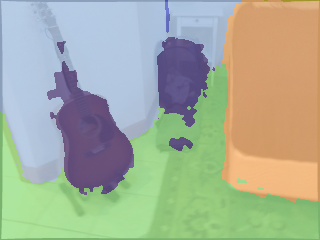}}&
\raisebox{-.4\height}{\includegraphics[width=\ww]{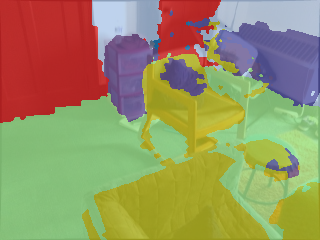}}&
\raisebox{-.4\height}{\includegraphics[width=\ww]{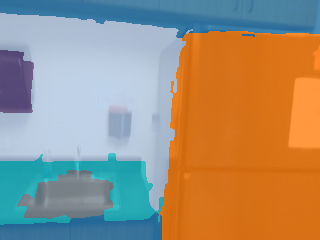}}&
\raisebox{-.4\height}{\includegraphics[width=\ww]{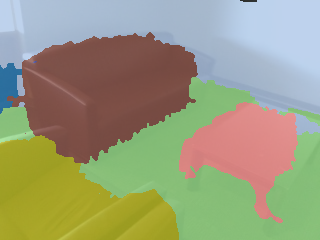}}\\[0.55cm]

{\small$\mathbf{B}_\texttt{G}$}&
\raisebox{-.4\height}{\includegraphics[width=\ww]{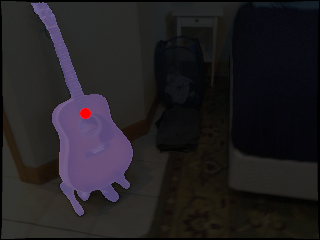}}&
\raisebox{-.4\height}{\includegraphics[width=\ww]{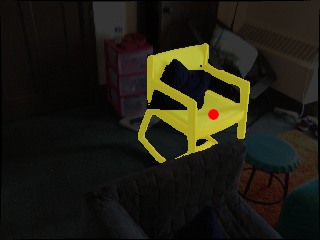}}&
\raisebox{-.4\height}{\includegraphics[width=\ww]{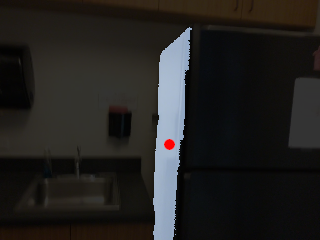}}&
\raisebox{-.4\height}{\includegraphics[width=\ww]{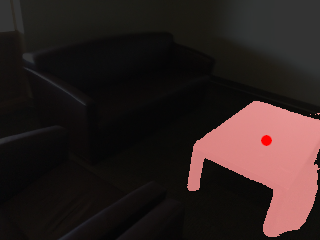}}\\[0.55cm]

{\small$\mathbf{Y}_\texttt{G}^\texttt{IR}$}&
\raisebox{-.4\height}{\includegraphics[width=\ww]{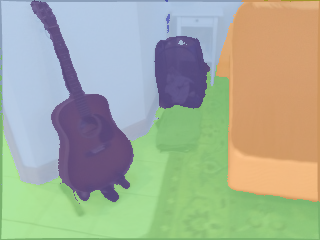}}&
\raisebox{-.4\height}{\includegraphics[width=\ww]{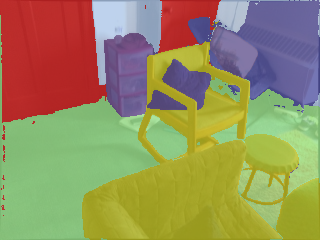}}&
\raisebox{-.4\height}{\includegraphics[width=\ww]{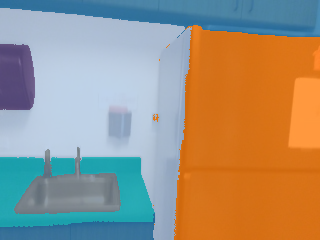}}&
\raisebox{-.4\height}{\includegraphics[width=\ww]{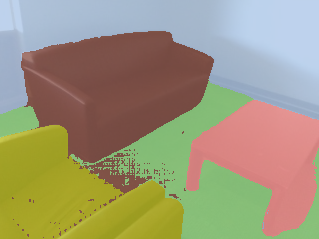}}\\[0.55cm]
    
{\small$\mathbf{B}_\texttt{I}$}&

\raisebox{-.4\height}{\includegraphics[width=\ww]{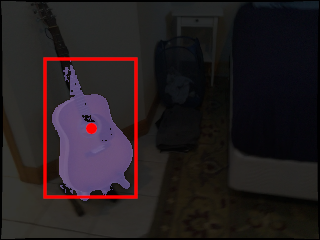}}&
\raisebox{-.4\height}{\includegraphics[width=\ww]{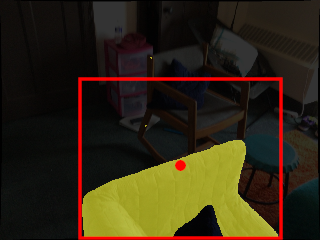}}&
\raisebox{-.4\height}{\includegraphics[width=\ww]{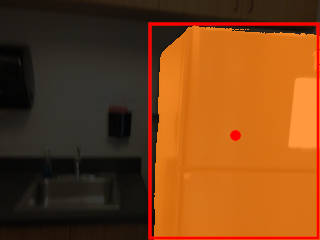}}&
\raisebox{-.4\height}{\includegraphics[width=\ww]{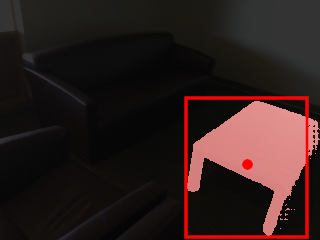}}\\[0.55cm]

{\small$\mathbf{Y}^\texttt{IR}_\texttt{I}$ }&

\raisebox{-.4\height}{\includegraphics[width=\ww]{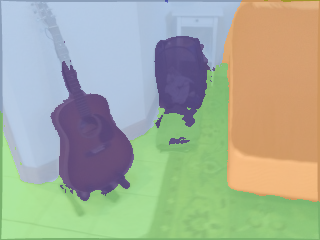}}&
\raisebox{-.4\height}{\includegraphics[width=\ww]{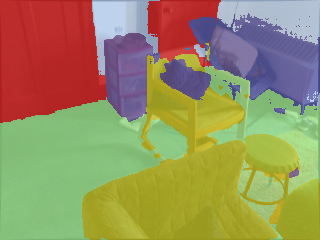}}&
\raisebox{-.4\height}{\includegraphics[width=\ww]{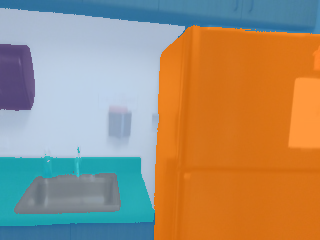}}&
\raisebox{-.4\height}{\includegraphics[width=\ww]{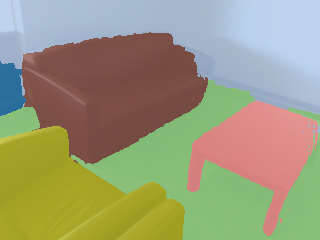}}\\

\end{tabular}
    %\caption{Comparison of the pseudo--labels obtained with our method using the \texttt{grid} ($\mathbf{Y}^\texttt{IR}_\texttt{G}$) and the \texttt{informed} ($\mathbf{Y}^\texttt{IR}_\texttt{I}$) prompt strategies, together with a representative object instance extracted from SAM and its relative prompt ($\mathbf{B}_\texttt{G}$ and $\mathbf{B}_\texttt{I}$ for \texttt{grid} and \texttt{informed}, respectively). $\mathbf{I}^t$ and $\mathbf{Y}^\texttt{MC}$ are the RGB images and the multi--view consistent annotations. In the first two columns, the \texttt{grid} approach ensures better results, while in the last two we can see how guiding the prompts with \texttt{informed} increases the precision of the pseudo--labels.}
    \caption{Comparison of pseudo--labels from our method using \texttt{grid} ($\mathbf{Y}^\texttt{IR}_\texttt{G}$) and \texttt{informed} ($\mathbf{Y}^\texttt{IR}_\texttt{I}$) prompts, with representative SAM instances and their prompts ($\mathbf{B}_\texttt{G}$, $\mathbf{B}_\texttt{I}$). $\mathbf{I}^t$ and $\mathbf{Y}^\texttt{MC}$ denote the RGB images and multi--view annotations. The first two columns highlight the strengths of \texttt{grid}, while the last two show the improved precision of \texttt{informed}.}
    \label{fig:promts_diff}
    \vspace{-0.7cm}
\end{figure}

\setlength{\tabcolsep}{0.43em}
\begin{table*}[!htbp]
\caption{Performance of the adapted model where training and testing use different sequences of the same environment}
\vspace{-0.2cm}
\scriptsize
\centering

\begin{tabular}{c|c|cc|cccccc||c|cc|cccccc}
\toprule
&&\multicolumn{2}{c|}{Baseline (5cm)} & \multicolumn{6}{c||}{Our method}&&\multicolumn{2}{c|}{Baseline (3cm)} & \multicolumn{6}{c}{Our method}\\
Env & \PT & \MC& $\Delta_{\text{\PT}}$ &\IRG & $\Delta_{\text{\MC}} $& \IRI & $\Delta_{\text{\MC}} $ & \IRGI & $\Delta_{\text{\MC}} $& \PT& \MC& $\Delta_{\text{\PT}}$ &\IRG & $\Delta_{\text{\MC}} $& \IRI & $\Delta_{\text{\MC}} $ & \IRGI & $\Delta_{\text{\MC}} $ \\
\midrule
1&	47,1&	49,6&	5,3\%&      \bestvalue{55}&	    10,9\%&	51,2&	3,2\%& \secondvalue{54,3}&	9,5\%	& 47,1&	48,9&	3,8\%&	    \secondvalue{50,6}&	3,5\%&	    49,5&	1,2\% &\bestvalue{50,8}	&3,9\%\\[0.05cm]
2&	\bestvalue{44,1}&	35,2&	-20,2\%&    39,6&	12,5\%&	\secondvalue{39,6}&	12,5\%& 39,3&	11,6\%&	 \bestvalue{44,1}&	\secondvalue{42,1}&	-4,5\%&	    41,8&	-0,7\%&	      41,6&	-1,2\%&41,6 &	-1,2\%\\[0.05cm]
3&	30,6&	30,5&	-0,3\%&	    32,8&	7,5\%&   \secondvalue{34,4}&	12,8\%&	  \bestvalue{34,7} &	13,8\%& 30,6&	31,5&	2,9\%&	    \bestvalue{36,5}&	15,9\%&	     35,1&	11,4\%&\secondvalue{36,3}&	15,2\%\\[0.05cm]
4&	55,3&	55,8&	0,9\%&      \secondvalue{57,2}&	2,5\%&   56,2&	0,7\%& \bestvalue{57,7}	&3,4\%&    55,3&	57,1&	3,3\%&	    \bestvalue{57,4}&	0,5\%&	    56,8&	-0,5\%&\secondvalue{57,2}&	0,2\%\\[0.05cm]
6&	49,2&	50,5&	2,6\%&	   52,8&	4,6\%&  \bestvalue{54,7}&	8,3\%& \secondvalue{53}&	5,0\%&   49,2&	51,2&	4,1\%&	    54,8&	7,0\%&	    \bestvalue{55,8}&	9,0\%&\secondvalue{54,8}	&7,0\%\\[0.05cm]
7&	52&	54,4&	4,6\%&	55,5&	2\%&	\bestvalue{57,2}&	    5,1\%& \secondvalue{56,2}	&3,3\%&   52&	54,1&	4\%&54,8&	1,3\%&	    \secondvalue{55,1}&	1,8\%&\bestvalue{55,3}&	2,2\%\\[0.05cm]
10&	57,5&	63,2&	9,9\%&	    \secondvalue{68,7}&	8,7\%&  67,6&	7,0\%& \bestvalue{68,7}&	8,7\%&   57,5&	65,2&	13,4\%&	    \secondvalue{69,1}&	6,0\%&	    68,3&	4,8\%&\bestvalue{70,1}&	7,5\%\\\midrule
Avg&48&	48,5&	0,4\%&	    \secondvalue{51,7}&	7\%&  51,6&	7,1\%&  \bestvalue{52,0}&	7,9\%&  48&	50&	3,9	\%&    \secondvalue{52,1}&	4,8	\%&    51,7&	3,8\%&\bestvalue{52,3}	&5,0\%\\

\bottomrule 
\multicolumn{19}{l}{}\\[-0.2cm]
\multicolumn{19}{p{0.9\textwidth}}{mIoU (\bestvalue{best} and \secondvalue{second best} values) of $f_{\theta_t}^{seg}$ trained with our method (\IRG, \IRI, and \IRGI) against the baseline (\MC), and the pre--trained model (\PT).}
\end{tabular}

%\caption{Improvements of $f_{\theta_t}^{seg}$ trained using the instance--aware pseudo--labels using \texttt{grid} (\IRG), \texttt{informed} (\IRI), and the combination (\IRGI) of the strategies compared with the fine--tuning using the multi--view consistent annotations (\MC) and the per--frame outputs of the pre--trained model $f_{\theta_s}^{seg}$ (\PT). The voxel size is set to 5cm and 3cm. Training and testing are performed using different sequences of the same environment. Best and second best performances are in \bestvalue{bold} and \secondvalue{underlined}.}
\label{tab:performance_fine_tune_100}
\vspace{-0.3cm}

\end{table*}
\renewcommand{\imageone}{voxel_5/scene0006_01/1946.jpg}
\renewcommand{\imagetwo}{voxel_5/scene0009_01/0.jpg}
\renewcommand{\imagethree}{voxel_5/scene0002_01/4084.jpg}
\renewcommand{\imagefour}{voxel_3/scene0000_01/4775.jpg}
\renewcommand{\imagefive}{voxel_3/scene0003_01/1438.jpg}

\renewcommand{\ww}{0.13\linewidth}
\begin{figure*}[!htbp]
\begin{tabular}{@{}c@{ }c@{ }c@{ }c@{ }c@{ }c@{ }c@{ }}
{\small $\mathbf{I}^t$} & {\small \PT} &{\small \MC} & {\small \IRG} & {\small \IRI} & {\small \IRGI} & {\small $\mathbf{Y}^\texttt{GT}$} \\
\includegraphics[trim=0 0 1920 0, clip, width=\ww]{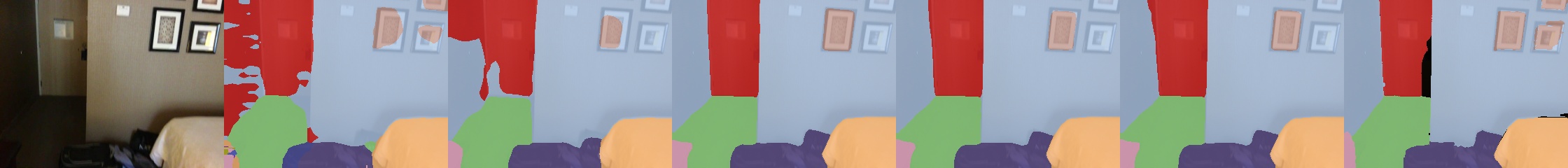}
&\includegraphics[trim=320 0 1600 0, clip, width=\ww]{contents/images/deeplab/\imageone}
&\includegraphics[trim=640 0 1280 0, clip, width=\ww]{contents/images/deeplab/\imageone}
&\includegraphics[trim=960 0 960 0, clip, width=\ww]{contents/images/deeplab/\imageone}
&\includegraphics[trim=1280 0 640 0, clip, width=\ww]{contents/images/deeplab/\imageone}
&\includegraphics[trim=1600 0 320 0, clip, width=\ww]{contents/images/deeplab/\imageone}
&\includegraphics[trim=1920 0 0 0, clip, width=\ww]{contents/images/deeplab/\imageone}\\

\includegraphics[trim=0 0 1920 0, clip, width=\ww]{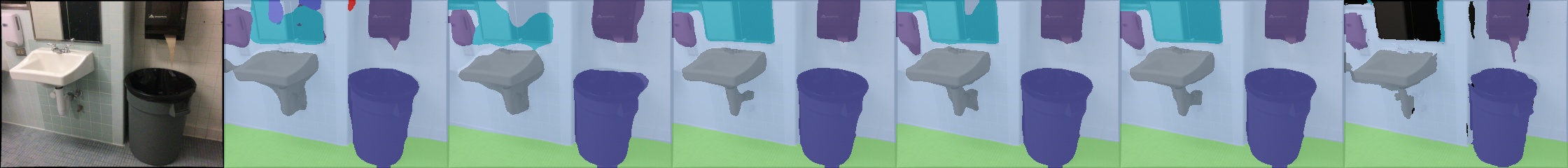}
&\includegraphics[trim=320 0 1600 0, clip, width=\ww]{contents/images/deeplab/\imagetwo}
&\includegraphics[trim=640 0 1280 0, clip, width=\ww]{contents/images/deeplab/\imagetwo}
&\includegraphics[trim=960 0 960 0, clip, width=\ww]{contents/images/deeplab/\imagetwo}
&\includegraphics[trim=1280 0 640 0, clip, width=\ww]{contents/images/deeplab/\imagetwo}
&\includegraphics[trim=1600 0 320 0, clip, width=\ww]{contents/images/deeplab/\imagetwo}
&\includegraphics[trim=1920 0 0 0, clip, width=\ww]{contents/images/deeplab/\imagetwo}\\

\begin{comment}
\includegraphics[trim=0 0 1920 0, clip, width=\ww]{contents/images/deeplab/\imagethree}
&\includegraphics[trim=320 0 1600 0, clip, width=\ww]{contents/images/deeplab/\imagethree}
&\includegraphics[trim=640 0 1280 0, clip, width=\ww]{contents/images/deeplab/\imagethree}
&\includegraphics[trim=960 0 960 0, clip, width=\ww]{contents/images/deeplab/\imagethree}
&\includegraphics[trim=1280 0 640 0, clip, width=\ww]{contents/images/deeplab/\imagethree}
&\includegraphics[trim=1600 0 320 0, clip, width=\ww]{contents/images/deeplab/\imagethree}
&\includegraphics[trim=1920 0 0 0, clip, width=\ww]{contents/images/deeplab/\imagethree}\\
\end{comment}

\includegraphics[trim=0 0 1920 0, clip, width=\ww]{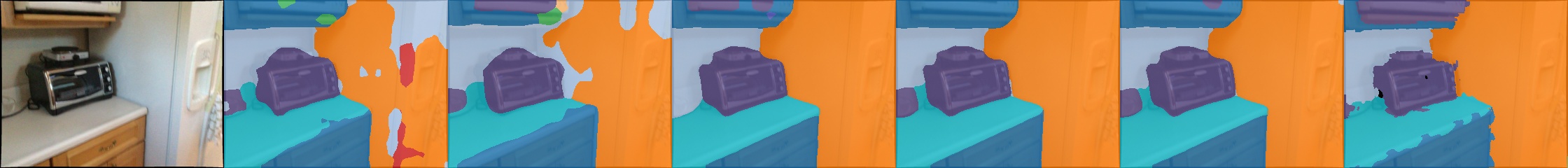}
&\includegraphics[trim=320 0 1600 0, clip, width=\ww]{contents/images/deeplab/\imagefour}
&\includegraphics[trim=640 0 1280 0, clip, width=\ww]{contents/images/deeplab/\imagefour}
&\includegraphics[trim=960 0 960 0, clip, width=\ww]{contents/images/deeplab/\imagefour}
&\includegraphics[trim=1280 0 640 0, clip, width=\ww]{contents/images/deeplab/\imagefour}
&\includegraphics[trim=1600 0 320 0, clip, width=\ww]{contents/images/deeplab/\imagefour}
&\includegraphics[trim=1920 0 0 0, clip, width=\ww]{contents/images/deeplab/\imagefour}\\

\includegraphics[trim=0 0 1920 0, clip, width=\ww]{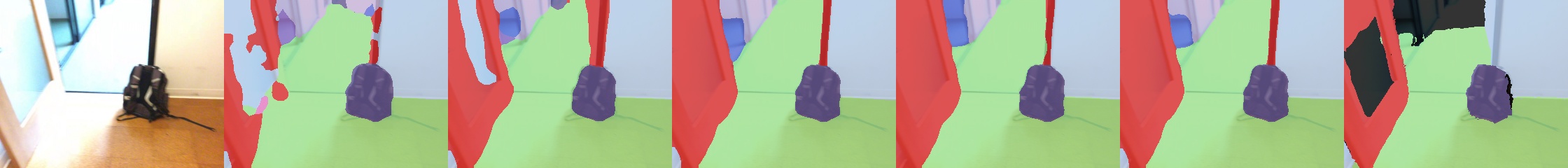}
&\includegraphics[trim=320 0 1600 0, clip, width=\ww]{contents/images/deeplab/\imagefive}
&\includegraphics[trim=640 0 1280 0, clip, width=\ww]{contents/images/deeplab/\imagefive}
&\includegraphics[trim=960 0 960 0, clip, width=\ww]{contents/images/deeplab/\imagefive}
&\includegraphics[trim=1280 0 640 0, clip, width=\ww]{contents/images/deeplab/\imagefive}
&\includegraphics[trim=1600 0 320 0, clip, width=\ww]{contents/images/deeplab/\imagefive}
&\includegraphics[trim=1920 0 0 0, clip, width=\ww]{contents/images/deeplab/\imagefive}\\

\end{tabular}
\centering
\begin{tabular}{llllllll}

\classwall&\classfloor&\classrefrigerator&\classcabinet&\classcounter&\classbed&\classdoor&\classsofa\\\classsink&\classtable&\classchair&\classmirror&\classwindow&\classpicture&\classotherfurniture&\classotherprop

\begin{comment}
\classwall
&\classfloor
&\classcabinet
&\classbed
&\classchair
&\classsofa\\
\classtable
&\classdoor
&\classwindow
&\classbookshelf
&\classpicture
&\classcounter\\
\classblinds
&\classdesk
&\classshelves
&\classcurtain
&\classdresser
&\classpillow\\
\classmirror
&\classfloormat
&\classclothes
&\classceiling
&\classbooks
&\classrefrigerator\\
\classtelevision
&\classpaper
&\classtowel
&\classshowercurtain
&\classbox
&\classwhiteboard\\
\classperson
&\classnightstand
&\classtoilet
&\classsink
&\classlamp
&\classbathtub\\
\classbag
&\classotherstructure
&\classotherfurniture
&\classotherprop
&
\end{comment}

\end{tabular}

    %\caption{Examples of the segmentation improvements (obtained in a new sequence) provided by our method using different prompting strategies (\IRG, \IRI, and \IRGI) with respect to the baselines of \MC{} and \PT. Images of the first (last) two rows are obtained by setting the voxel size to 5cm (3cm). $\mathbf{I}^t$ and $\mathbf{Y}^\texttt{GT}$ are the RGB images and the ground--truth annotations.}
    \caption{Segmentation improvements with different prompting strategies (\IRG, \IRI, \IRGI) compared to the baselines (\MC{} and \PT). The first two rows use a $5\text{cm}$ voxel size, the last two a $3\text{cm}$. %$\mathbf{I}^t$ and $\mathbf{Y}^\texttt{GT}$ denote the RGB images and ground--truth annotations.
    Black areas in $\mathbf{Y}^\texttt{GT}$ are due to rendering errors or missing labels; these segmentation errors are improved by our IR.
    }
    \label{fig:deeplab_fine_tune_new_sequence}
    \vspace{-0.6cm}
\end{figure*}
In contrast, \texttt{informed} is a more conservative approach because the prompts are extracted from the multi--view consistent annotations. This is particularly useful to prevent the over--segmentation, a well--known limitation of SAM that often splits a single object into multiple instances, especially when the prompt is particularly dense (as in the case of \texttt{grid}). See, for example, the third column of Fig.~\ref{fig:promts_diff}, where \texttt{informed} identifies the fridge as a unique object, while \texttt{grid} considers its left side as a separate instance, which is then wrongly labeled. Another important effect of \texttt{informed} is to guide the predictions of SAM when objects have similar textures, a situation in which an automatic segmentation is particularly challenging. This is illustrated in the last column of Fig.~\ref{fig:promts_diff}, where \texttt{informed} identifies improved and more precise instances than \texttt{grid}, which partially mixes the table and the couches with the floor.

%\input{contents/images/figures/promts_differences}

%\subsection{Unsupervised domain adaptation results}
\subsection{Performance of the adapted model}
In this section, we present the experimental results evaluating the performance of our method on the semantic segmentation task. Table~\ref{tab:performance_20} shows the results obtained when fine--tuning and testing inside the same sequence, using the first $80\%$ and the last $20\%$ of the frames, respectively (as done in~\cite{continual_adaptation_2d3d} and~\cite{uda_neural_rendering}). We run all the experiments setting the voxel size of Kimera Semantics to $3\text{cm}$ and $5\text{cm}$. 

The results demonstrate that our instance--aware refinement remarkably increases the model's performance with respect to the multi--view consistency baseline of~\cite{continual_adaptation_2d3d} (\MC), and the raw predictions of the pre--trained $f^{seg}_{\theta_s}$ (\PT). Considering a voxel size of $5\text{cm}$, our method improves the baseline \MC{} by $4,3\%$ and $5,7\%$ using the \texttt{grid} (\IRG) and \texttt{informed} (\IRI) prompt strategies. As observed in the previous section, the impact of the multi--view consistency on the pre--trained model \PT{} varies a lot across environments, registering remarkable improvements (e.g., in environments $1$, $7$, and~$9$), but also important degradations (such as in environments $2$, $6$, and~$8$). Starting from this, our method improves the performance of \MC{} in both cases: the instance--aware refinement step not only increases the segmentation accuracy when \MC{} is effective, but also mitigates and fixes errors of the multi--view consistency in case of failure. See, for example, scenes $2$ and $6$: while \MC{} degrades \PT{} by  $\approx$ -14\% and $-9\%$, our method \IRI{} substantially improves the quality of the baseline $\approx$ 10\% and 7\%, obtaining segmentation accuracies close to \PT. 
Table~\ref{tab:performance_20} also shows the benefits of fine--tuning the source model $f^{seg}_{\theta_s}$ combining the pseudo--labels obtained with the \texttt{grid} and \texttt{informed} prompting strategies (\IRGI). Mixing the two different types of instance--aware annotations during training mitigates the negative intrinsic effects of \texttt{grid} and \texttt{informed} (previously described in Section~\ref{sec:pseudo_labels_improvement}), helping the model in providing more accurate outputs. More specifically, \IRGI{} reaches an average increment of $6,7\%$ over the baseline \MC, achieving the best or the second best performance across all environments (except for scene $7$ where \IRI{} and \MC{} perform better). All these outcomes are confirmed using a voxel size of $3\text{cm}$. As expected, the quality of the baseline \MC{} is improved from $44,9$ to $46,3$ mIoU. This is because a smaller voxel size limits the discretization of the world encoded in the 3D map, thus reducing the artifacts produced by the ray--tracing procedure. Despite this, our method increases the performance of the baseline \MC{} by $3,6\%$ and $3,2\%$ with \IRG{} and \IRI{}, respectively. Again, the best performances are obtained by mixing the prompting strategies (\IRGI) during training, with an average increment of $4,4\%$ over \MC. Interestingly, while the baseline \MC{} is strongly influenced by the voxel size of Kimera Semantics, our method achieves comparable performance regardless of the voxels' resolution ($0.6$ mIoU difference obtained \IRGI{} with $3\text{cm}$ and $5\text{cm}$).  This is particularly useful in practical deployment scenarios, where the voxel dimension can be set to a higher value, thus reducing the computational overhead of the robot in integrating the frames in Kimera Semantics.

While the aforementioned results of Table~\ref{tab:performance_20} are collected in a subportion of the environment, Table~\ref{tab:performance_fine_tune_100} reports the performance of $f^{seg}_{\theta_t}$ considering the entire environment, as training and testing are performed in different sequences. %This evaluation protocol closely reflects the deployment of a real robotic platform, which is adapted using an initial data acquisition, and then tested in a different run. 
The results of this additional experiment further strengthen the findings of the previous evaluation.  
Using a voxel size of $5\text{cm}$ ($3\text{cm}$), our method outperforms the \MC{} baseline on average by $\approx$ 7\% ($4\%$), obtaining a remarkable performance increment in all scenes. The only case where the pre--trained model (\PT) performs worse after adaptation is in scene $2$, which is a problematic environment since almost the $30\%$ of the frames have missing positions and a large portion of the scene is not annotated. Despite this, our instance refinement strongly increase the baseline's performance of $\approx$ 12\% with respect to \MC. 
Again, the best averaging performances are obtained by \IRGI, which reaches almost the same performance using $5\text{cm}$ and $3\text{cm}$ as voxel sizes, respectively $52$ and $52,3$ mIoU. 
The improvements provided by our method can be observed in the qualitative examples of Fig.~\ref{fig:deeplab_fine_tune_new_sequence}. Despite the baseline (\MC) effectively removes the noise of the raw model's predictions (\PT), the $f^{seg}_{\theta_t}$ learns the errors that are persistent errors between multiple frames, like the doors in the first and last rows that are confused with the wall. In contrast, our method remarkably increases the quality of the segmentation masks produced by $f^{seg}_{\theta_t}$, solving also some artifacts of the ground--truth. This can be seen by observing the precision of the masks related to the sink and the furniture in the second and third rows. 

\section{Conclusions}

This paper presents a novel approach for unsupervised domain adaptation in robotics applications based on self--supervision with instance--aware pseudo--labels. In future investigations, we plan to embed the instance segmentation inside the 3D map and leverage foundation models to fix the semantic class of object instances. 

%\section*{Acknowledgements}
%Part of this project was funded by National Plan for NRRP Complementary Investments (PNC) in the call for the funding of research initiatives for technologies and innovative trajectories in the health---project n. PNC0000003--AdvaNced Technologies for Human-centrEd Medicine (project acronym: ANTHEM).
\bibliographystyle{IEEEtran}
\bibliography{contents/references}

\end{document}